\documentclass[lettersize,journal]{IEEEtran}
\usepackage{amsmath,amsfonts}
\usepackage[noend]{algpseudocode}
\usepackage{booktabs}
\usepackage{indentfirst}
\usepackage{threeparttable}
\usepackage{algorithmicx}%第二代排版
\usepackage{algorithm}
\usepackage{amsmath}
\usepackage{array}
\usepackage{multirow}
\usepackage{textcomp}
\usepackage{stfloats}
\usepackage{url}
\usepackage{tabularx}
\usepackage{verbatim}
\usepackage{graphicx}
{}  
\usepackage{float}
\usepackage{booktabs} 
\usepackage{subfigure}
\usepackage{cite}
\usepackage{colortbl}
\usepackage{CJKutf8}
\usepackage{hyperref} %生成引用链接，注：该宏包可能与其他宏包冲突，故放在所有引用的宏包之后
\usepackage{cleveref}
\definecolor{mygreen}{rgb}{.85,1,.85}

\begin{document}
\begin{CJK}{UTF8}{gbsn}

\title{Fuzzy-NMS: Improving 3D Object Detection with Fuzzy Classification in NMS}

\author{Li Wang, Xinyu Zhang, Fachuan Zhao, Chuze Wu, Yichen Wang, Ziying Song, Lei Yang, Jun Li, Huaping Liu
        % <-this % stops a space
\thanks{This work was supported by the National High Technology Research and Development Program of China under Grant No. 2018YFE0204300, the National Natural Science Foundation of China under Grant No. 62273198, U1964203, the China Postdoctoral Science Foundation (No.2021M691780), and State Key Laboratory of Robotics and Systems (HIT) (SKLRS-2022-KF-12). (Corresponding author: Xinyu Zhang.)}% <-this % stops a space

\thanks{L. Wang is with the State Key Laboratory of Automotive Safety and Energy, and the School of Vehicle and Mobility, Tsinghua University, Beijing 100084, and also with State Key Laboratory of Robotics and Systems (HIT), Harbin 150001, China (e-mail: wangli\_thu@mail.tsinghua.edu.cn).}

\thanks{X. Zhang, F. Zhao, C. Wu, Y. Wang, L. Yang and J. Li are with the State Key Laboratory of Automotive Safety and Energy, and the School of Vehicle and Mobility, Tsinghua University, Beijing 100084, China (e-mail: xyzhang@tsinghua.edu.cn; chofachuan-@163.com; wcz1999@sjtu.edu.cn; 22S136042@stu.hit.edu.cn; yanglei20@ mails.tsinghua.edu.cn; lj19580324@126.com).}

\thanks{Ziying Song is with the School of Computer and Information Technology, Beijing Key Lab of Traffic Data Analysis and Mining, Beijing Jiaotong University, Beijing 100044, China (e-mail: 22110110@bjtu.edu.cn).}

\thanks{H. Liu is with the State Key Laboratory of Intelligent Technology and the Systems and Department of Computer Science and Technology, Tsinghua University, Beijing 100084, China (e-mail: hpliu@tsinghua.edu.cn).}}

\markboth{Journal of \LaTeX\ Class Files,~Vol.~14, No.~8, August~2021}%
{Shell \MakeLowercase{\textit{et al.}}: A Sample Article Using IEEEtran.cls for IEEE Journals}


\maketitle

\begin{abstract}
Non-maximum suppression (NMS) is an essential post-processing module used in many 3D object detection frameworks to remove overlapping candidate bounding boxes. However, an overreliance on classification scores and difficulties in determining appropriate thresholds can affect the resulting accuracy directly. To address these issues, we introduce fuzzy learning into NMS and propose a novel generalized Fuzzy-NMS module to achieve finer candidate bounding box filtering. The proposed Fuzzy-NMS module combines the volume and clustering density of candidate bounding boxes, refining them with a fuzzy classification method and optimizing the appropriate suppression thresholds to reduce uncertainty in the NMS process. Adequate validation experiments are conducted using the mainstream KITTI \textcolor{black}{and large-scale Waymo} 3D object detection benchmarks. The results of these tests demonstrate the proposed Fuzzy-NMS module can improve the accuracy of numerous recently NMS-based detectors significantly, \textcolor{black}{including PointPillars, PV-RCNN, and IA-SSD, etc.} This effect is particularly evident for small objects such as pedestrians and bicycles. As a plug-and-play module, Fuzzy-NMS does not need to be retrained and produces no obvious increases in inference time.
\end{abstract}


% figure1
\begin{figure}[!t]
	\vspace{-5mm}
	\centering
 	\includegraphics[width=1\linewidth]{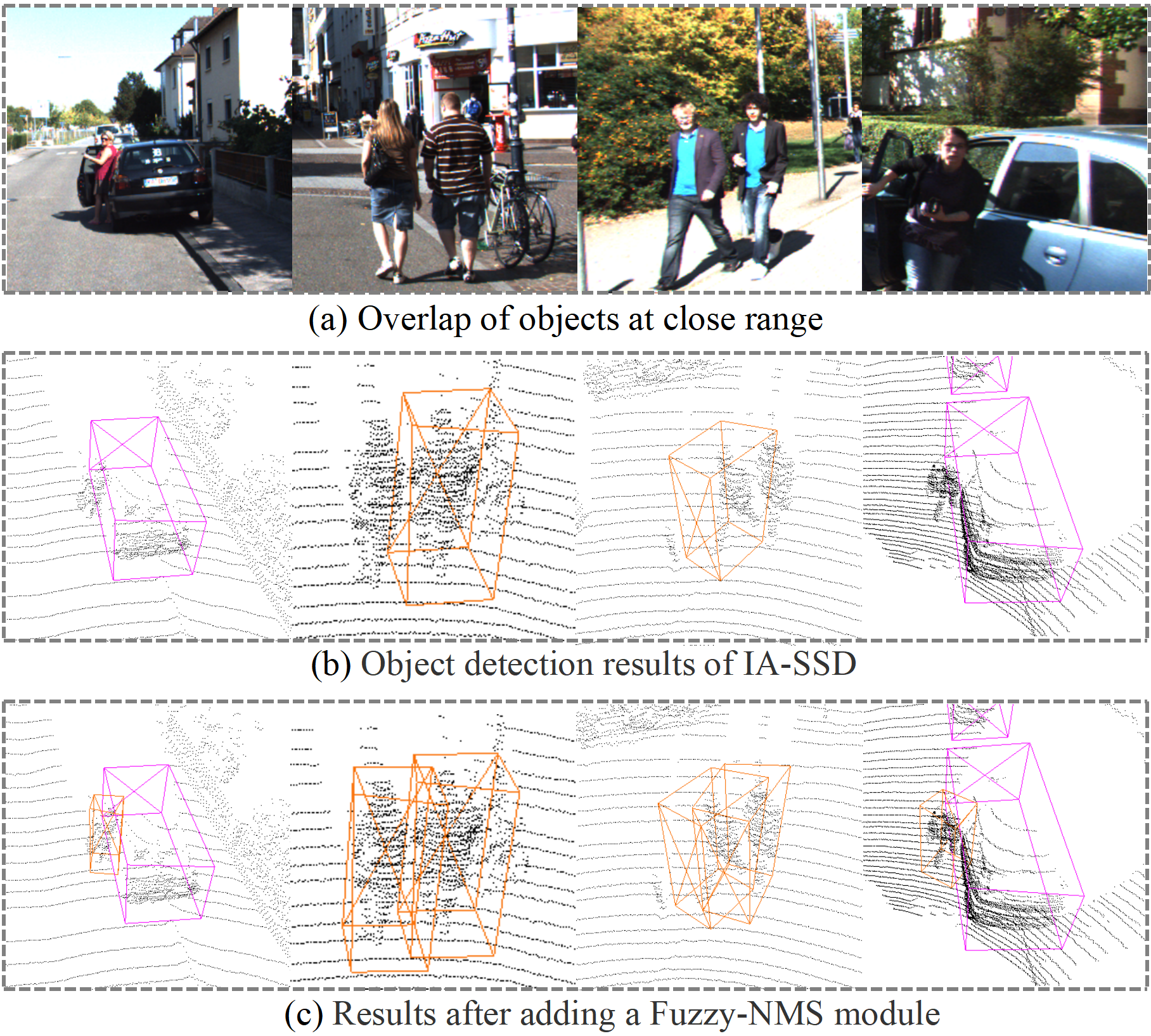}
	\caption{\textcolor{black}{Comparative experiments on adjacent object detection. Subfigure (a) shows the overlap of objects at close range, and subfigures (b) and (c) show the 3D object detection results of IA-SSD\cite{zhang2022not} without and with our proposed Fuzzy-NMS module, respectively. Experiments illustrate that our algorithm can effectively reduce the missed detection of adjacent objects. 
		\vspace{-4mm}}}
	\label{fig1}
\end{figure}

\section{Introduction}
% Motivation: 
% 分类score和定位置信度不一致，错误的类别检测，体积

\IEEEPARstart
%{T}{hree} dimensional (3D) object detection has become essential for autonomous driving. 
{T}{he} \textcolor{black}{prominence of 3D vision has been increasingly recognized, encompassing tasks such as 3D segmentation \cite{yang2022nested} and 3D detection, with 3D object detection in particular emerging as a pivotal aspect for autonomous driving.} Unlike 2D object detection, 3D object detection can provide an object category, spatial location, 3D size, and movement direction simultaneously, which is conducive to subsequent planning and vehicle control \cite{qi2018frustum}. Thus, 3D object detection is more practical yet more challenging. 

As a common post-processing module, non-maximum suppression (NMS) has received extensive attention in object detection studies. Like 2D object detectors \cite{shafiee2017fast, girshick2015fast}, 3D object detectors can also be divided into NMS-based and NMS-free categories, determined by the presence or absence of an NMS module. NMS-based methods, such as PointPillars \cite{lang2019pointpillars}, SECOND \cite{yan2018second}, and \textcolor{black}{BSAODet \cite{xiao2023balanced}}, offer the advantage of more stable network training by including the anchor size as a priority. Dense anchor boxes can then effectively improve network recall capabilities, which produces obvious benefits for small object detection \cite{chen2020hierarchical, li2020rtm3d, li2019gs3d}. \textcolor{black}{Compared to the previous PointNet-like modules, GMPNet \cite{yin2021graph} effectively encourages information dissemination among different grids while capturing short-term motion clues of objects. The above-mentioned works all use the traditional NMS algorithm for post-processing, and their fixed NMS threshold also brings many obvious disadvantages. Compared with these methods, the selection of the NMS threshold in our method is more flexible and can be more efficient. It works well for scenes with different densities and different numbers of objects. }  NMS-free methods, such as Fa3d \cite{demilew2020fa3d}, AFDET \cite{ge2020afdet}, and Centerpoint\cite{yin2021center}, exhibit the greatest advantages in terms of detection speed. Since there is no need for preset anchors, only the object center point, length, width, and height of feature maps need to be regressed at different scales, which greatly reduces the runtime. The primary disadvantage is that the accuracy is typically lower than NMS-based methods, which are more widely used to balance accuracy with operational efficiency.

Furthermore, NMS is often used to preliminarily filter bounding boxes predicted in the detection head\cite{ye2020monocular}. Inaccurate intersection over union (IoU) thresholds can often result from differing categories, scenes, scales, and detected object proportions, which can seriously affect network performance. This is especially problematic for point cloud data, in which large objects (e.g., cars and trucks) are often far apart from each other in the real world, and the corresponding IoU threshold is lower \cite{meyer2019lasernet}. In contrast, small objects (e.g., pedestrians and bicycles) can be very close in reality, and the corresponding IoU threshold is higher. Thus, candidate bounding boxes can be processed using the mathematical characteristics of 3D boxes to acquire optimal IoU thresholds \cite{ fan2021rangedet}. However, due to the changing environment and objects, the determination of optimal preprocessing and threshold values remains challenging.

Previous observations have suggested overlapping phenomenons between objects in space, as shown in \mbox{Fig. \ref{fig1}}. \textcolor{black}{For example, in  \mbox{Fig. \ref{fig1}}(a), we show two different scenarios, namely: pedestrians walking side by side and pedestrians leaning against the car, both of which will lead to overlap between the same category or between different categories. As shown in  \mbox{Fig. \ref{fig1}}(b), when using a recent 3D object detection method named IA-SSD\cite{zhang2022not} with a traditional NMS, two pedestrians walking side by side are mistakenly detected as one pedestrian because they are very close together, and the pedestrian next to the car is not detected. In other words, when objects are closer together, their density increases, resulting in a potentially more significant overlap. However, the strength of this overlap can vary due to the degree of association between objects of different volumes, such as pedestrians or cars. Additionally, extensive annotation can further enlarge the connections between such spaces. Different objects have varying sizes and degrees of overlap, so when performing post-processing with NMS, corresponding thresholds need to be provided. In \mbox{Fig. \ref{fig1}}(c), we also show the visualization effect after adding the Fuzzy-NMS module for comparison. It can be seen that pedestrians walking side by side are distinguished, and the passersby leaning against the car are also detected.}

Based on this observation, we model the prediction of volume and distribution density for bounding boxes as a mathematical characteristic for use in determining object spatial correlation. This information can provide strong prior knowledge for guiding candidate box classification in the detection process and is utilized to develop a fuzzy classification system designed to adjust the optimal NMS threshold. Specifically, all bounding boxes preliminarily predicted by the detector head are divided into three categories based on 3D characteristics, which include large-volume high-density (LVHD), small-volume high-density (SVHD), and low-density (LD). This classification is achieved by introducing the DBSCAN clustering algorithm, which is used first to estimate the density of initially predicted bounding boxes and then calculate the volume of these boxes\cite{birant2007st,hahsler2019dbscan}. \textcolor{black}{Given a set of points in some space, it marks points that are closely packed together (with many neighbors) as outliers, which are located in low-density regions where the nearest neighbors are too far away. DBSCAN \cite{schubert2017dbscan} is one of the most common clustering methods and the most cited algorithm in the leading data mining works\cite{shen2016real}.} Generally, we observe that higher densities produce larger NMS thresholds for small objects, allowing different thresholds to be assigned to different categories. By classifying the initially predicted bounding boxes and assigning different NMS thresholds, an optimal value can be easily selected. However, the definitions of density and volume in 3D classification tasks exhibit great uncertainty as a result, which can seriously affect detection performance. To this end, a fuzzy classification system named Fuzzy-NMS is constructed to assist each bounding box in determining the optimal NMS threshold. This fuzzy model takes the density and volume of candidate boxes as input, applies a triangular membership function to fuzzify the input, and then uses prior knowledge to establish fuzzy rules between feature vectors and corresponding categories. This method exhibits strong robustness and produces accurate classification results for different types of bounding boxes. In this process, each bounding box is assigned to a different category using fuzzy classification, and an accurate threshold is specified. The predicted scores are utilized for the final bounding box screening in the Fuzzy-NMS module. The primary contributions of this work can be summarized as follows:
%Previous statements have demonstrated that fuzzy classification can effectively classify samples with significant uncertainty. 
\begin{enumerate}
\item{We introduce a novel fuzzy classification system (Fuzzy-NMS) for 3D object detection. In this process, deep learning and fuzzy classification are combined to establish a novel paradigm for 3D object detection.}
\item{Human prior knowledge is introduced into the NMS post-processing module, effectively associating the volume and clustering density of candidate bounding boxes with the distribution of objects in space, which is implemented to optimize appropriate suppression thresholds for various categories, thereby reducing uncertainty while removing overlapping boxes.}
\item{The proposed Fuzzy-NMS module has been proven to be effective for numerous existing 3D object detection algorithms, \textcolor{black}{including PointPillars\cite{lang2019pointpillars}, PV-RCNN\cite{shi2020pv}, Part-A2-Anchor\cite{shi2020points} and IA-SSD\cite{zhang2022not}, etc.} Specifically, assigning different NMS thresholds to various bounding boxes in the KITTI dataset\cite{geiger2013vision} and \textcolor{black}{Waymo open dataset\cite{sun2020scalability}} produces significant performance improvements in multiple categories (cars, pedestrians, and cyclists).}
\end{enumerate}

% The remainder of this paper is organized as follows. Section II describes related work in 3D object detection, non-maximum suppression, and fuzzy classification performed in recent years. Section III introduces the detailed Fuzzy-NMS structure. Section IV describes the dataset and experimental setups used to conduct comparative experiments. Finally, conclusions and future work are provided in Section V.

\section{related work}
% This section briefly reviews related studies on 3D object detection, non-maximum suppression, and fuzzy classification systems.

\subsection{3D Object Detection}
Currently, deep learning-based 3D object detectors are mainly divided into Voxel-based and Point-based methods based on their processing of point clouds. The NMS module plays a crucial role in 3D detectors \cite{vpnet}. % STD \cite{zhang2015single} involves a point-based proposal generation paradigm used for object detection in point clouds with spherical anchors, utilizing the NMS based on a classification score and targeted BEV IoUs to eliminate redundant results during proposal generation.
\textcolor{black}{3D-VID\cite{yin2020lidar} takes advantage of the time-continuous relationship between frames that have not been used by their predecessors, the author adds grap-based GNN convolution on the basis of PointPillars, which expands the receptive field of each node. NMS with an IOU threshold of 0.5 is utilized when generating the final detections. ProposalContrast \cite{yin2022proposalcontrast} proposes a novel framework for unsupervised point cloud pre-training to learn robust 3D representations via contrastive region proposals. Yin et al. propose a cluster-based box voting (CBV) module, which groups seed boxes into different clusters and aggregates votes for each seed box in the cluster to generate a more exact box. In this way, cleaner and more accurate pseudo-labels can be obtained by a simple NMS\cite{yin2022semi}. HCPVF \cite{fan2023hcpvf} proposes a new BEV attention method to learn more distinct point cloud features, through two cascaded linear layers and a normalization layer to mine the point similarity in BEV and reduce the uneven sampling of sparse BEV features. \cite{tao2023pseudo} proposes a new pseudo-monocular 3D object detection algorithm, which learns object-aware features through feature enhancement and guided query initialization, which reduces the high computational cost brought by the use of additional data. In recent years, NMS-free methods like \cite{li2021one} have also occupied an important position in the field of 3D object detection, which does not need to analyze the sample distribution of the data set to obtain the optimal setting of the prior anchor box. Nevertheless, the NMS-free method still has a certain gap in accuracy compared with the NMS-based method.} These studies suggest that object detection using an NMS module could be a practical tool for 3D object detection.

\subsection{Non-Maximum Suppression}
\textcolor{black}{Non-Maximum Suppression (NMS) is an integral part of many deep learning methods and is widely used in object detection, object tracking, 3D reconstruction, and texture analysis \cite{homma2021non}. The core idea of NMS is to find local maxima and suppress non-maximum. It effectively eliminates duplicate detection boxes and keeps the box with the highest confidence \cite{huang2020nms, salscheider2021featurenms, choi2021standard}. Traditional NMS methods are often too strict in eliminating redundant boxes, and the elimination mechanism based only on the IoU threshold is not good for occlusion cases \cite{he2018softer}. To solve this problem, Soft-NMS  \cite{bodla2017soft} and Adaptive-NMS \cite{liu2019adaptive} are proposed. Soft-NMS \cite{bodla2017soft} uses a penalty mechanism to reduce the confidence of the detection frame whose IoU is greater than the threshold, instead of directly setting it to zero. Adaptive NMS \cite{liu2019adaptive} dynamically adjusts the threshold according to the object aggregation and mutual occlusion, so as to realize the adaptive adjustment threshold.  Similar to our idea, Decoupled R-CNN \cite{wang2022decoupled} has noticed the influence of fixed NMS threshold on a high-density proposal in the field of two-dimensional target detection, so the corresponding NMS classification threshold and regression are dynamically changed in the experiment to explore the effect of NMS threshold on two different tasks.  There are also some works dedicated to the improvement of NMS, such as OS-NMS proposed by \cite{meng2021towards}. Unlike our method, which changes the threshold, this work changes the center point selection strategy of the pre-selection box, assuming that a point with a higher foreground score is more accurate and close to the center, and tends to make a more confident center prediction. And then, for each center point with a high score, the distance from it in other pre-selected centers is less than a certain value. In addition, \cite{sun2022drone} proposes IA-NMS, which multiplies the classification of the visible light mode by the corresponding alignment coefficient to reduce the occurrence of false alarms. Unlike our method, it mainly improves the detection results in dark scenes. NMS has also been applied to 3D object detection, such as distance-guided NMS \cite{qi2018frustum} and mathematically differentiable NMS framework with backpropagation \cite{wang2021salient}, for improving the performance of distance estimation and monocular 3D images, respectively. Center-Aware Non-Maximum Suppression (CA-NMS) \cite{meng2020weakly} introduces the idea that points near the center are easier to predict the center than points far from the center, and assigns higher foreground scores to points near the center through a pseudo-label generation strategy. These studies show that the performance of NMS in different classes and overlapping situations can be further improved by directly or indirectly choosing an appropriate threshold.}

\subsection{Fuzzy System}
\textcolor{black}{A fuzzy system is a model in which the input, output, and state variables are defined based on fuzzy sets. Fuzzy systems can imitate comprehensive human inference in dealing with fuzzy information problems that are difficult to solve with conventional mathematical methods \cite{siminski2022prototype, wu2020fuzzy}. This approach can also be used to solve non-linear problems and has been widely applied to pattern recognition, decision analysis, and medical diagnosis \cite{zheng2021design}. Fuzzy systems have also been introduced to assist in salient object detection \cite{bechini2020tsf}. For example, Nozaki proposes an object suggestion selection and measurement mechanism involving fuzzy set theory and uses the acquired scores to detect salient objects \cite{nozaki1994selecting}. Fuzzy systems have also been used for segmentation tasks \cite{ishibuchi1998fuzzy, phong2009classification, bai2017symmetry}.  Although fuzzy methods have applications and effects in many tasks, they are still rarely applied to 3D object detection, especially in NMS. Therefore, to filter candidate bounding boxes more precisely, we introduce fuzzy classification into NMS and apply prior knowledge to help eliminate overlapping candidate bounding boxes. Some prior knowledge can be expressed by using a fuzzy classification model to subdivide candidate boxes, optimizing appropriate suppression thresholds for various categories, thus effectively reducing uncertainty in the NMS process. }

\begin{figure*}[!t]
	\vspace{-5mm}
	\centering
	\includegraphics[width=\linewidth,height=7.2cm]{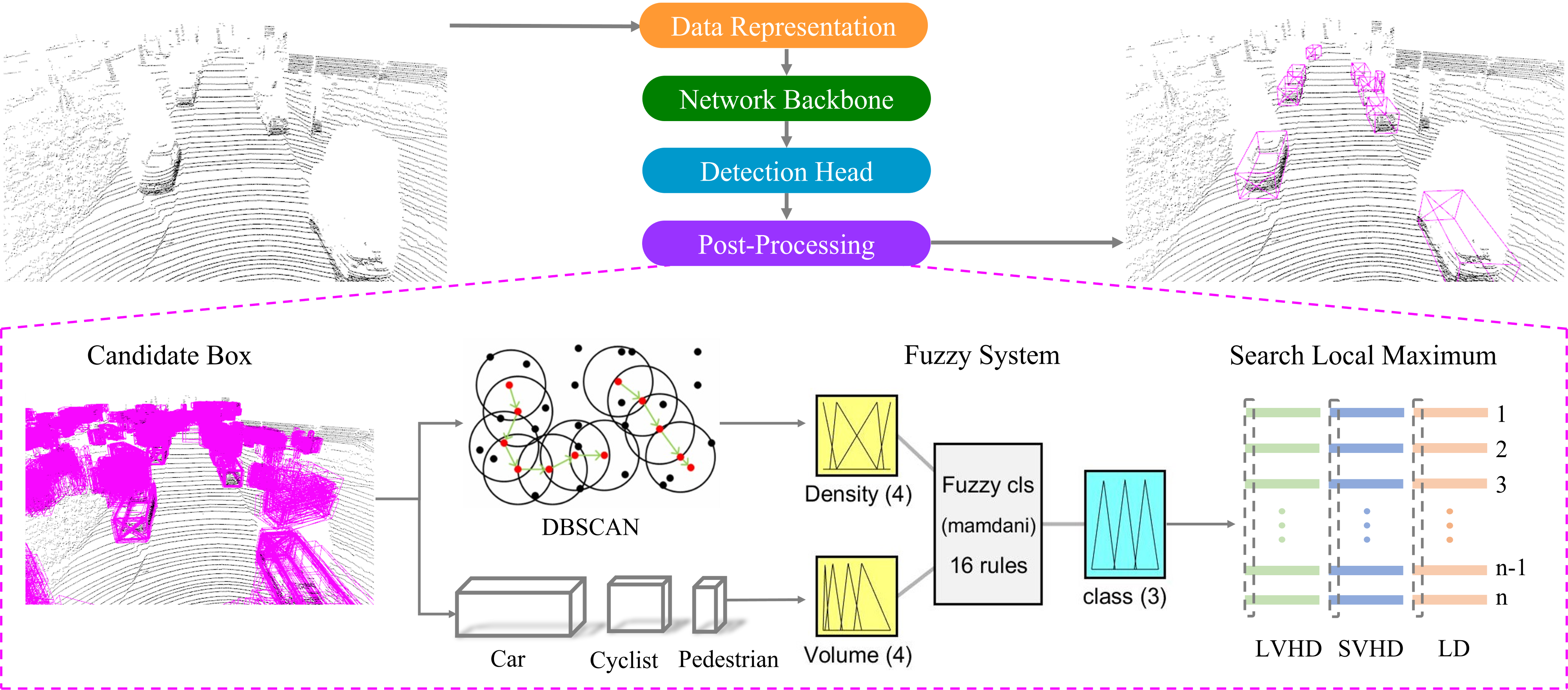}
	\caption{The 3D object detection framework with our proposed Fuzzy-NMS. The entire network can be divided into pre-3D detection and post-processing components. The input point cloud is processed in data representation, network backbone, and detection head steps used to regress candidate bounding boxes, which can be performed using existing 3D object detection methods. The proposed Fuzzy-NMS module is applied during post-processing. We calculate the volumes of bounding boxes after confidence filtering and the corresponding densities using DBSCAN clustering. Bounding boxes are then categorized by fuzzy classification. Various thresholds are then set, and local maximums are searched, which effectively eliminates redundant detection boxes.
		\vspace{-4mm}}
	\label{fig2}
\end{figure*}

\section{METHOD}
\subsection{Overview}
Existing 3D object detection networks can be divided into 3D data pre-processing, feature extractor, detection head, and post-processing modules. In this study, we mainly improve the post-processing component of the network. In other words, the proposed Fuzzy-NMS is a plug-and-play module compatible with most existing 3D object detection networks. Specifically, as shown in \mbox{Fig. \ref{fig2}}, the input from the LiDAR point cloud is first pre-processed and then sent to the feature extractor and detection head to identify preliminary bounding boxes. The density and volume of these preliminary boxes can be estimated using the DBSCAN cluster algorithm. The proposed Fuzzy-NMS then uses these values to classify each bounding box through a fuzzy classification system intended to reduce uncertainty. Assigning different thresholds to various bounding box categories improves the resulting detection performance. Finally, bounding boxes are further filtered using scores acquired in the detection head. Fuzzy-NMS has been shown to be an effective improvement over many existing 3D object detection methods.

\subsection{3D Bounding Box Classification}
\begin{figure}[t]
\subfigure[3D view]{
\includegraphics[width=4.
cm,height = 2.8cm]{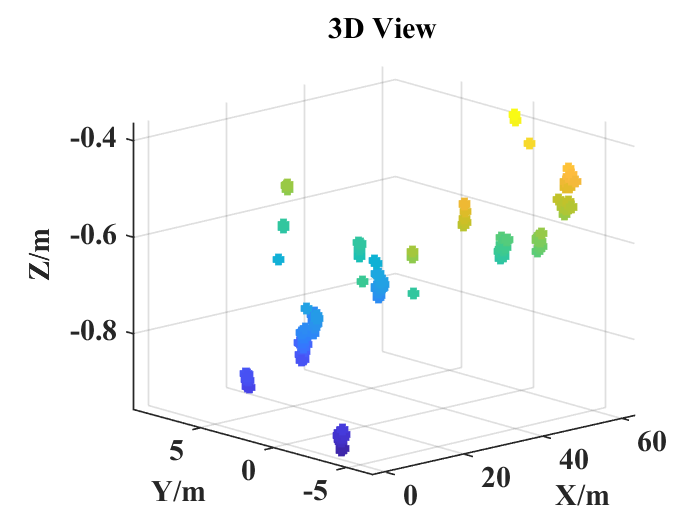}}
\subfigure[Front view]{
\includegraphics[width=4.2cm,height = 2.8cm]{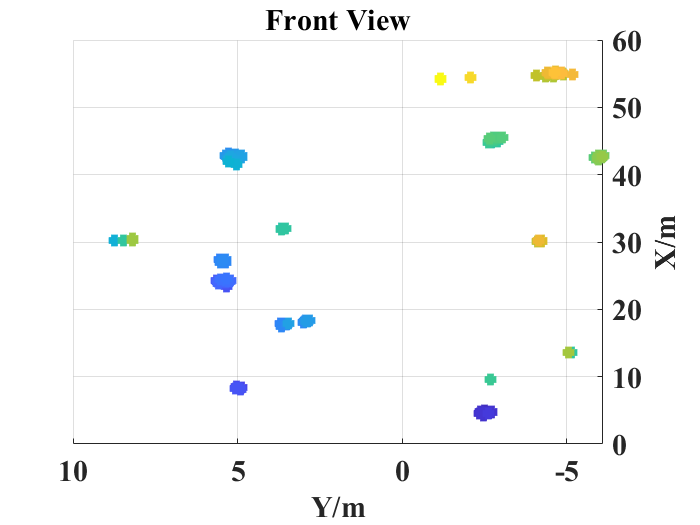}}
\caption{The distribution of candidate box center points in space. The first set of point cloud data (000000.bin) in the KITTI 3D object detection training set is used to illustrate the problem.}
\label{fig3}
\end{figure}

NMS threshold selection has been shown to have an important impact on object detection performance. For example, in the KITTI dataset\cite{geiger2013vision}, if PointPillars is allocated an NMS threshold value of 0.01 (for initial bounding box prediction), it cannot identify the best threshold value for all bounding boxes with varying volumes and densities. The result is a detection box that cannot effectively eliminate duplicates. For this reason, it is necessary to assign NMS thresholds using customized criteria. This study makes full use of the mathematical characteristics of 3D information and includes these data as prior knowledge for dividing the predicted bounding boxes into three categories. As mentioned, when a bounding box is sparsely distributed, a small NMS threshold should be used to eliminate redundant boxes. In contrast, a higher threshold should be used when boxes are densely distributed to achieve a higher precision. \mbox{Fig. \ref{fig3}} demonstrates this connection between correlation density and detection accuracy, as isolated boxes are mostly misclassified (box centers are shown in the figure). Compared with higher-density candidate boxes, there is a greater possibility for lower-density boxes to be incorrectly selected. 

However, in this study, predicted boxes are divided into three categories: large-volume high-density (LVHD), small-volume high-density (SVHD), and low-density (LD). In the case of small boxes, the number of anchors to be matched is much lower than that of larger objects, as most boxes near the anchors are redundant. In this case, a higher IoU threshold should be set to eliminate redundant boxes. The above analysis indicates that the density and volume of bounding boxes determine the optimal IoU threshold to a certain extent. In other words, since low-density boxes are generally sparse, dividing their volumes into two categories is not usually advantageous. Therefore, we just regard them as a low-density (LD) category.

\subsection{Volume and Density Estimation}
As discussed previously, Fuzzy-NMS divides candidate bounding boxes into LVHD, SVHD, and LD categories based on density and volume, which should be calculated prior to classification. Volume can be determined as:
\begin{equation}
d_{v_{i}}=d_{x_{i}} \times d_{y_{i}} \times d_{z_{i}}(i=1,2,3 \cdots),
\end{equation}
where $d_{x_{i}}, d_{y_{i}}, d_{z_{i}}$ represent the length, width, and height of the $i-th$ predicted box, respectively.

Density calculation requires the use of a clustering algorithm, which clusters the candidate boxes, measures the density of each cluster, and then uses this value to determine the density of individual boxes. Specifically, the DBSCAN algorithm is selected for this step, as it can cluster dense datasets of any shape and does not require specifying the number of clusters. In addition, it offers fast speed and effective noise processing, so it is often used for spatial data. DBSCAN categorizes points into specific categories, including core, boundary, and noise points. After noise removal, edges are established to connect core points that are less than one domain radius from the core point search space (denoted Eps). These connected core points are then formed into a cluster, and boundary points are finally assigned to the corresponding cluster radius. In this study, the center coordinates of candidate bounding boxes are used as input, and the density clustering of adjacent boxes is achieved using the DBSCAN algorithm. Box density can be expressed as follows:

\begin{equation}
D_{k}=\frac{N_{k}}{\max \left(N_{0}, N_{1} \ldots N_{k}\right)},
\end{equation}
where \textit{k} represents the \textit{k}-th cluster, $D_{k}$ is the density of the \textit{k}-th cluster, and $N_{k}$ is the number of bounding boxes contained in the \textit{k}-th cluster. It should be noted that $N_{0}$ represents noise points, which are typically sparse. Eq. (2) restricts the clustering of 3D bounding boxes to values between 0 and 1, to facilitate subsequent Fuzzy-NMS calculations.

\subsection{Fuzzy Classification System}
The class of fuzzy algorithms primarily includes pure fuzzy logic systems, Takagi-Sugeno fuzzy systems, and Mamdani fuzzy systems, which are most often used in practical engineering applications. Therefore, a Mamdani-type fuzzy system is used as the basic unit in this study. In this method, a fuzzy max-min synthesis operation is implemented as follows:

\begin{equation}
\mu_{B}(y)=\underset{x \subset X}{\vee}\left[\mu_{A}(x) \wedge \mu_{R}(x, y)\right],
\end{equation}
where $\vee$ represents the min, $\wedge$ represents the max, $\mathbf{A}$ is the set of $x$, $\mathbf{B}$ is the set of $y$, $\mathbf{R}$ is the set of $x$ and $y$ output relations, and $\mu_{A}(x)$, $\mu_{B}(y)$, and $\mu_{R}(x, y)$ denote membership values for the fuzzy sets $\mathbf{A}$, $\mathbf{B}$, and $\mathbf{R}$, respectively. 

In Fuzzy-NMS, the sum is selected as the output synthesis operation. Similarly, de-fuzzification involves the use of a centroid method that offers smoother output inference than alternatives. In other words, even if the input variable changes only slightly, the output will change considerably. This process can be represented as:

\begin{equation}
v_{O}=\frac{\int_{V} v \mu_{v}(v) d v}{\int_{V} \mu_{v}(v) d v},
\end{equation}
where $v_{O}$ represents the center of gravity for the area enclosed by the membership function curve and the abscissa, $µv$ is the membership function curve, and $v$ denotes a fuzzy input variable.

A fuzzy statistical method is adopted to determine the membership function for the input and output. Specifically, we fix the sample value $u$ and changed the set $A_{*}$. After $n$ experiments, the membership frequency ($P_{A}$) of $u$ through $A$ can be expressed as $P_{A}=\tau_{A} / n $, where $\tau_{A}$ indicates the number of times $u$ belongs to $A_{*}$). The membership function is determined to be triangular, expressible with the three parameters $a$, $b$, and $c$ as follows:

\begin{equation}
f(x, a, b, c)=\left\{\begin{array}{cc}
0 & x \leq a \\
\frac{x-a}{b-a} & a \leq x \leq b \\
\frac{c-b}{c-x} &  b \leq x \leq c\\
0 & x>c
\end{array}, \right.
\end{equation}
where $a$ and $c$ determine the ``foot" of the triangle, $b$ denotes the “peak” of the triangle, $x$ represents the input to the membership function, and  $f(x, a, b, c)$  defines the membership value. 

\begin{figure}[!t]
	%\vspace{-5mm}
	\centering
	\includegraphics[width=0.7\linewidth,height=4cm]{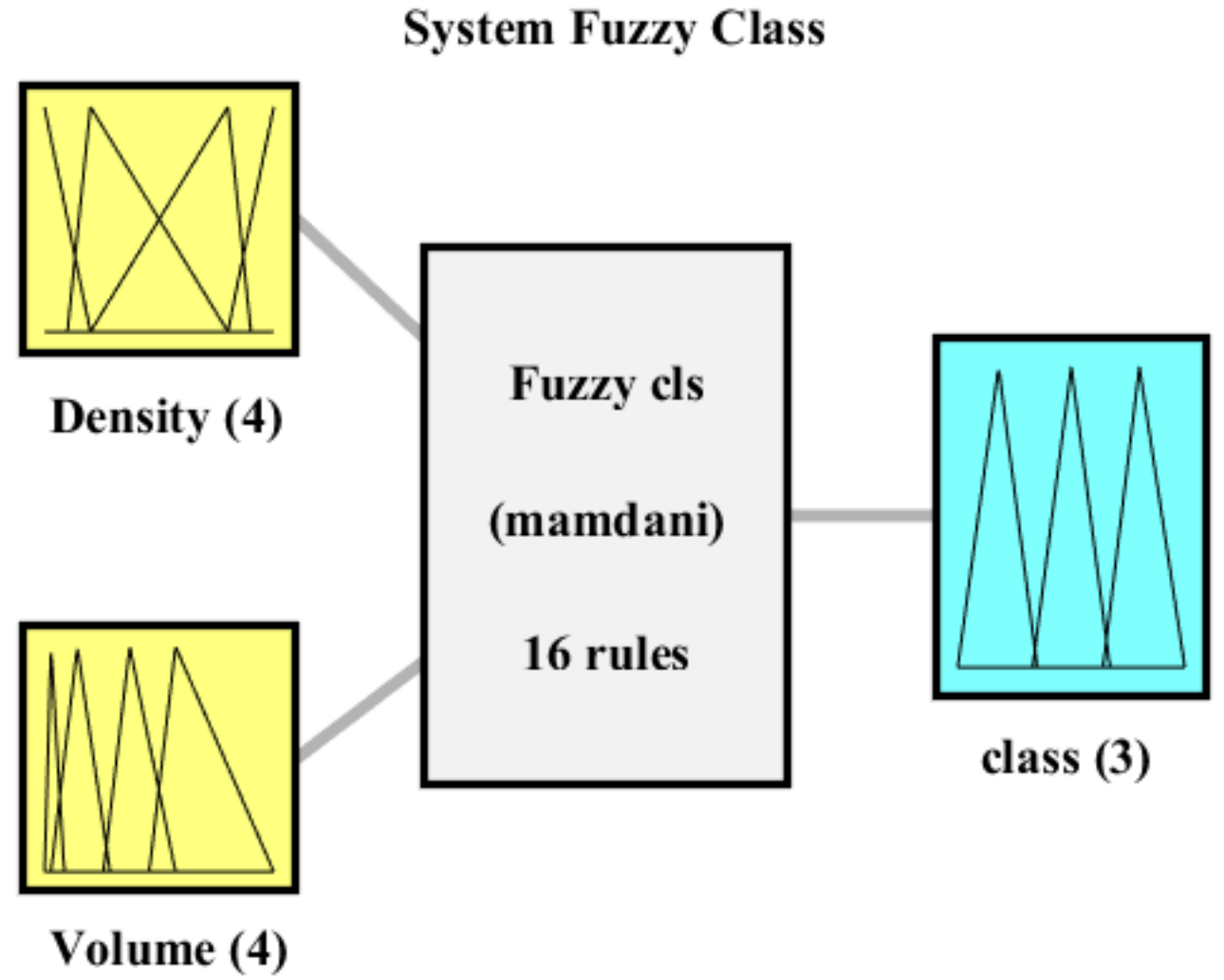}
	\caption{Mamdani fuzzy system created by fuzzy reasoning rules. The yellow part represents the input, the middle gray part represents the fuzzy rules, and the black part represents the output.}
	\label{fig4}
\end{figure}

Multi-dimensional fuzzy rules are then applied to construct a fuzzy rule library. A fuzzy rule table is also established, based on prior knowledge, to ensure the completeness of fuzzy rules, as shown in \mbox{Table \ref{table_1}}. Once the structure of the fuzzy classification system is determined, the framework is developed in the Matlab R2021A software platform, as shown in \mbox{Fig. \ref{fig4}}. The fuzzy system is then used to classify predicted bounding boxes by first counting the internal mathematical characteristics of the data as prior knowledge used for classification.

\begin{table}
\centering
\caption{FUZZY RULE TABLE}
\label{table_1}
\setlength{\tabcolsep}{6mm}
\renewcommand{\arraystretch}{1}%行高

\begin{tabular}{ccccc} 

\toprule[0.8pt]
\multirow{2}{*}{D} & \multicolumn{4}{c}{V}  \\ 
\cline{2-5}
                   & ZE & PS & PM & PB      \\ 
\hline
ZE                 & S  & S  & S  & S       \\
PS                 & S  & M  & B  & B       \\
PM                 & M  & M  & B  & B       \\
PB                 & M  & B  & B  & B       \\
\bottomrule[0.8pt]
\end{tabular}
\begin{tablenotes}
        \footnotesize
        \vspace{2mm}
        \item[*] \textcolor{black}{*Each symbol represents a fuzzy set, with subsets ZE, PS, PM, and PB representing zero, positive small, positive middle, and positive big. Subsets S, M, and B represent small, medium, and large.}  %此处加入注释*信息
\end{tablenotes}
\end{table}

\begin{figure}[!t]
    \vspace{-3mm}
    \centering
    \subfigure[Density statistics]{
    \includegraphics[width=4.2cm,height =3cm]{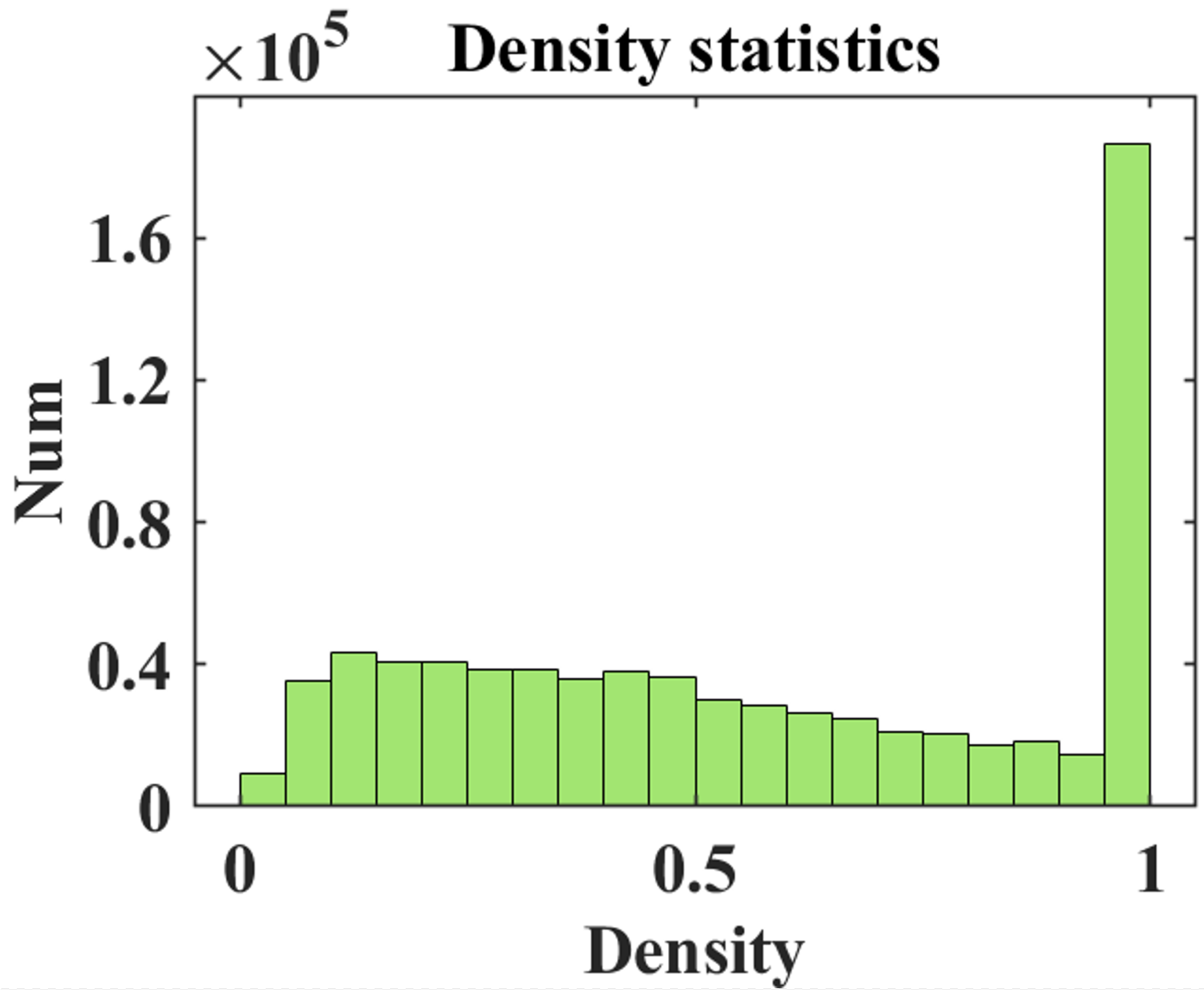}}
    \subfigure[Volume statistics]{
    \includegraphics[width=4.2cm,height = 3cm]{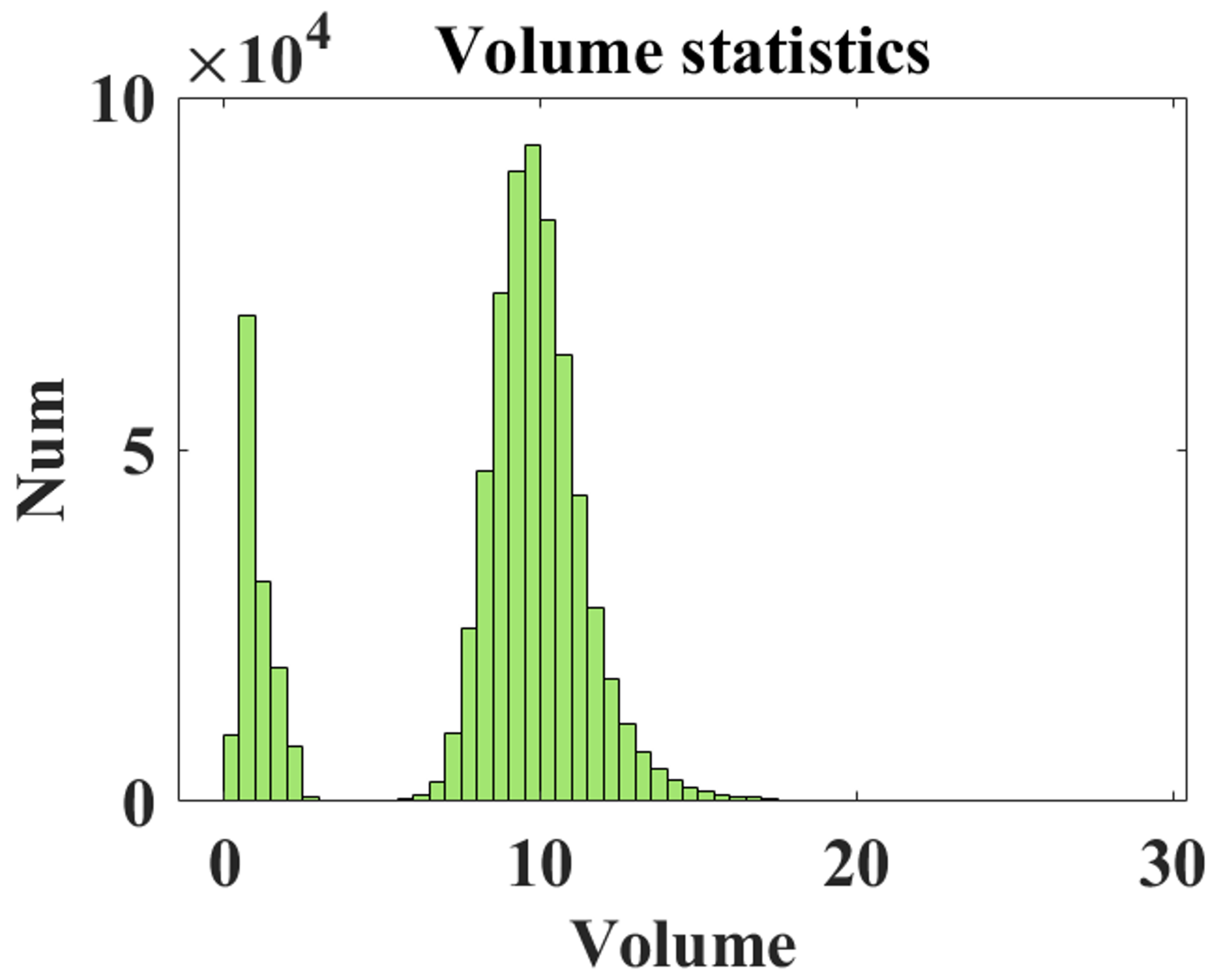}}

    \caption{Density and volume statistics for candidate boxes in the KITTI dataset. The first set of point cloud data (000000.bin) in the KITTI 3D object detection training set is used to illustrate the problem.
    \vspace{1mm}}
    \label{fig5}
\end{figure}

\mbox{Fig. \ref{fig5}} presents a statistical bar graph of the density and volume of candidate bounding boxes for a scene in the KITTI 3D object detection training set. It can be observed from the above statistical results that density is mostly concentrated between 0-1, while the volume is distributed between 0-15. As such, feature density and volume can be divided into four fuzzy sets: \textbf{ZE} (Zero), \textbf{PS} (Positive Small), \textbf{PM} (Positive Medium), and \textbf{PB} (Positive Big). The output variable categories can be divided into three fuzzy sets: \textbf{S} (Small), \textbf{M} (Medium), and \textbf{B} (Big). For the convenience of expression, \textbf{S} stands for \textbf{LD} (Low-Density), \textbf{M} stands for \textbf{SVHD} (Small-Volume High-Density), and \textbf{B} stands for \textbf{LVHD} (Large-Volume High-Density). A membership function is then assigned to each fuzzy set with a range of 0-1. Since the number of input variables is two, and each input variable includes four fuzzy sets, there are 16 fuzzy rules created based on prior knowledge. We then convert these fuzzy rules from \mbox{Table \ref{table_1}} into an IF-THEN ruleset, as shown in \mbox{Table \ref{table_2}}.

\begin{table}[]
\caption{IF-THEN RULESET}
\label{table_2}
\centering
\setlength{\tabcolsep}{1mm}
\renewcommand{\arraystretch}{1}
\begin{tabular}{ccccc}
% \toprule[0.8pt]
\cline{1-1}

\textbf{Fuzzy rule base : A collection of IF-THEN rules}           &  &  &  &  \\ \cline{1-1}
\begin{tabular}[c]{@{}l@{}}· If (density is ZE) and (volume is ZE) then (class is S) \\ · If (density is ZE) and (volume is PM) then (class is S) \\ · If (density is ZE) and (volume is PS) then (class is S) \\ · If (density is ZE) and (volume is PB) then (class is S) \\ · If (density is PS) and (volume is ZE) then (class is S) \\ · If (density is PS) and (volume is PM) then (class is M) \\ · If (density is PS) and (volume is PS) then (class is B) \\ · If (density is PS) and (volume is PB) then (class is B) \\ · If (density is PM) and (volume is ZE) then (class is M) \\ · If (density is PM) and (volume is PM) then (class is M) \\ · If (density is PM) and (volume is PS) then (class is B) \\ · If (density is PM) and (volume is PB) then (class is B) \\ · If (density is PB) and (volume is ZE) then (class is M) \\ · If (density is PB) and (volume is PM) then (class is B) \\ · If (density is PB) and (volume is PS) then (class is B) \\ · If (density is PB) and (volume is PB) then (class is B)\end{tabular} &  &  &  &  \\
\multicolumn{1}{c}{16 rules in total}                                         &  &  &  &  \\ 
% \bottomrule[0.8pt] 
\cline{1-1}

\end{tabular}
\end{table}

\subsection{Final Bounding Box Generation}
After constructing the fuzzy system, we classify different bounding boxes and assign thresholds for individual categories used to perform NMS. Finally, the scores produced by the detection head are applied to filter the predicted bounding boxes (after NMS), thereby acquiring the output result. The pseudo-code for this algorithm is shown in \cref{Algorithm I}.

\begin{algorithm}[t]
\caption{Fuzzy-NMS} 
\hspace*{0.00in}{\bf Input:}\\ %算法的输入， \hspace*{0.02in}用来控制位置，同时利用 \\ 进行换行
\hspace*{0.04in} {$\beta =\left \{ b_{1},...,b_{N} \right \},S =\left \{ s_{1},...,s_{N} \right \},N =\left \{ n_{1},...,n_{N} \right \}$}\\
\hspace*{0.04in} {$\beta$ is the list of initial detection boxes}\\
\hspace*{0.04in} {$S$ contains corresponding detection scores}\\
\hspace*{0.04in} {$N$ is the NMS threshold}\\
 {\bf Begin:} %算法的结果输出
\begin{algorithmic}[1]
\State $D =\left \{ d_{1},...,d_{N} \right \} \gets $ DBSCAN calculation box density% \State 后写一般语句
\State $V =\left \{ v_{1},...,v_{N} \right \} \gets $ Calculate box volume
\State $\left \{ \beta_{1}, \beta_{2}, \beta_{3} \right \}, \left \{ S_{1}, S_{2}, S_{3} \right \} \gets $ Fuzzy classification for D and V
\State $ E\gets \{ \}$
\For{i in [0,1,2]} % For 语句，需要和EndFor对应
    \While{$\beta_{i} \ne $ empty}
　　    \State $m\gets argmax(S_{i}) $
　　    \State $M\gets b_{m}$  
　　    \State $E\gets E\cup M; \beta_{i}\gets \beta_{i}-M$
　　    \For{$b_{i}$ in $\beta_{i}$}
　　        \If{IoU($M$, $b_{i}\ge n_{i}$)}
　　            \State $\beta_{i}\gets \beta_{i}-b_{i}; S_{i}\gets S_{i}-s_{i}$
　　        \EndIf
　　        \State \textbf{end}
        \EndFor
        \State \textbf{end}
    \EndWhile
    \State  \textbf{end}
\EndFor
\State  \textbf{end}

\State \Return E, S
\end{algorithmic}
\end{algorithm}

\section{EXPERIMENTS}
A series of experiments were conducted to demonstrate the proposed Fuzzy-NMS module could significantly improve the efficiency of 3D object detection. In this section, the dataset used in these experiments was introduced, and the experimental details and results were described.

\begin{table}[]
\centering
\caption{MEMBERSHIP FUNCTION PARAMETERS}
\label{table_3}
\setlength{\tabcolsep}{3.2mm}
\renewcommand{\arraystretch}{1}
\begin{tabular}{ccccc}
\toprule[0.8pt]
VARIABLE                 & FUZZY SET & A    & B    & C    \\ \hline
\multirow{4}{*}{DENSITY} & ZE        & 0.0  & 0.0  & 0.1  \\
                         & PS        & 0.1  & 0.2  & 0.5  \\
                         & PM        & 0.4  & 0.8  & 0.9  \\
                         & PB        & 0.9  & 1.0  & 1.0  \\ \hline
\multirow{4}{*}{VOLUME}  & ZE        & 0.0  & 0.0  & 3.0  \\
                         & PS        & 2.0  & 5.0  & 10.0 \\
                         & PM        & 9.0  & 12.0 & 20.0 \\
                         & PB        & 17.0 & 20.0 & 35.0 \\ \hline
\multirow{3}{*}{CLASS}   & S         & 0.0  & 0.25 & 0.35 \\
                         & M         & 0.34 & 0.5  & 0.65 \\
                         & B         & 0.64 & 0.85 & 1.0  \\ \bottomrule[0.8pt]
\end{tabular}
\end{table}

\begin{table}[]
\centering
\caption{FUZZY RULE TABLE}
\label{table_4}
\setlength{\tabcolsep}{6mm}
\renewcommand{\arraystretch}{1}
\begin{tabular}{cccc}
\toprule[0.8pt]
Threshold & LD & LVHD & SVHD \\
\hline
Score &0.1 & 0.1 & 0.3 \\
IoU & 0.01 & 0.6 & 0.0 \\ \bottomrule[0.8pt]
\end{tabular}
\end{table}
\vspace{-1.25em} 
\begin{figure}[t]
\subfigure[Density membership]{
\includegraphics[width=4.2cm,height =3.2cm]{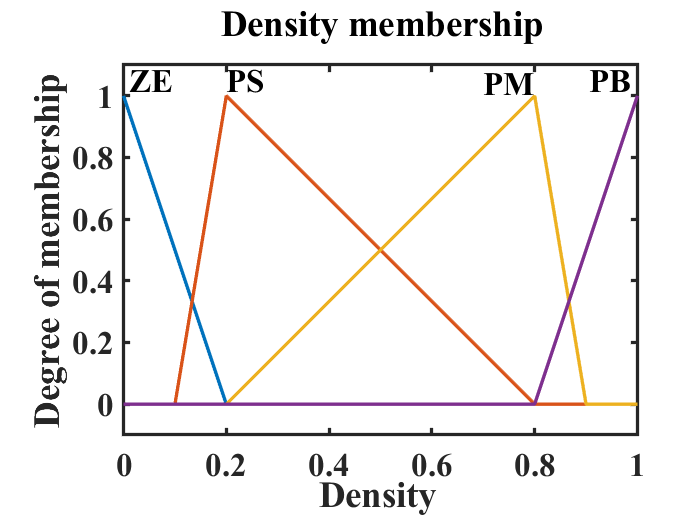}}
\subfigure[Volume membership]{
\includegraphics[width=4.2cm,height = 3.2cm]{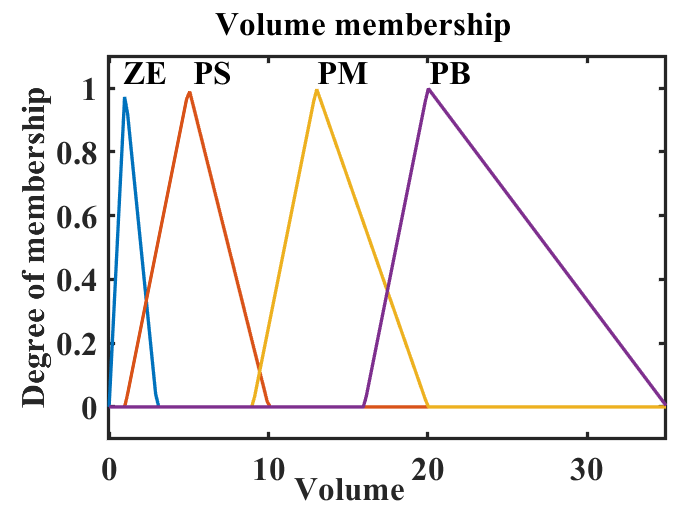}}
\centering
\subfigure[Classification membership]{
\includegraphics[width=4.2cm,height = 3.2cm]{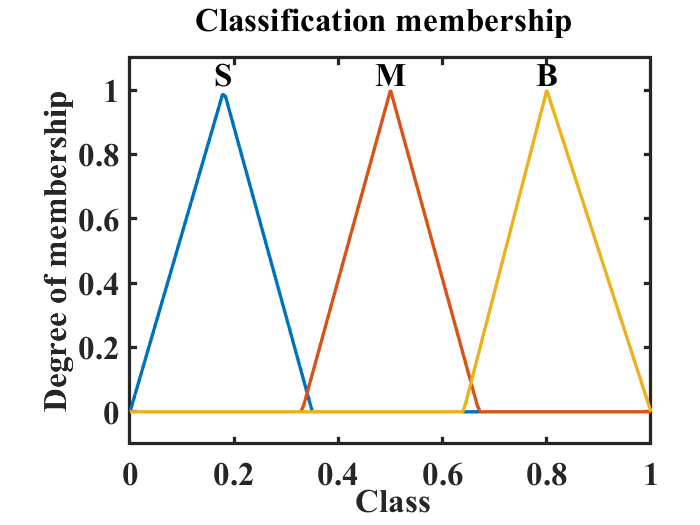}}
\caption{Input and output membership functions in the fuzzy system.}
\label{fig6}
\end{figure}

\subsection{Datasets and Evaluation Indicators}
The proposed method was evaluated using the KITTI 3D\cite{geiger2013vision} object detection data, which was currently the world’s most popular evaluation dataset for autonomous driving. The set contains 7,481 training samples and 7,518 test samples, primarily distributed across car, pedestrian, and cyclist categories. Detection can be divided into three difficulty levels: easy, moderate, and hard, depending on the degree of occlusion for marked bounding boxes. The mean average precision (mAP), the official evaluation standard for the KITTI 3D object detection benchmark, was used as an evaluation indicator in this study. \textcolor{black}{In addition, in order to prove the applicability of our method on large datasets, we also conducted verification experiments on Waymo Open Dataset\cite{sun2020scalability}, which is currently the largest dataset for 3D object detection of LiDAR point clouds in autonomous driving scenarios. This set contains 798 training sequences with around 160k LiDAR samples, 202 validation sequences with 40k LiDAR samples, and 150 testing sequences with 30k LiDAR samples. The evaluation of the system is conducted using official evaluation tools, employing mean average precision (mAP) and heading-weighted mean average precision (mAPH). The evaluation process involves two difficulty levels. In LEVEL 1, ground-truth objects are required to have a minimum of 5 inside points, while LEVEL 2 considers objects with at least 1 inside point as ground truth.}

\vspace{-1.25em} 
\begin{figure}[t]
\centering
\subfigure[DBSCAN clustering]{
\includegraphics[width=0.7\linewidth]{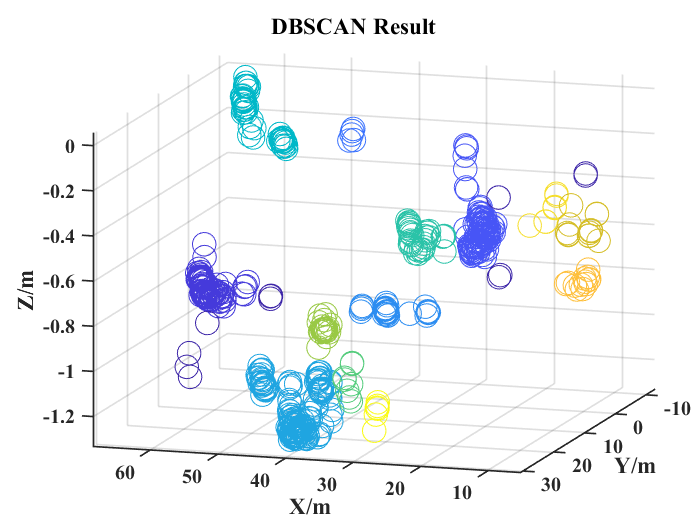}}
\centering
\subfigure[Fuzzy clustering]{
\includegraphics[width=0.7\linewidth]{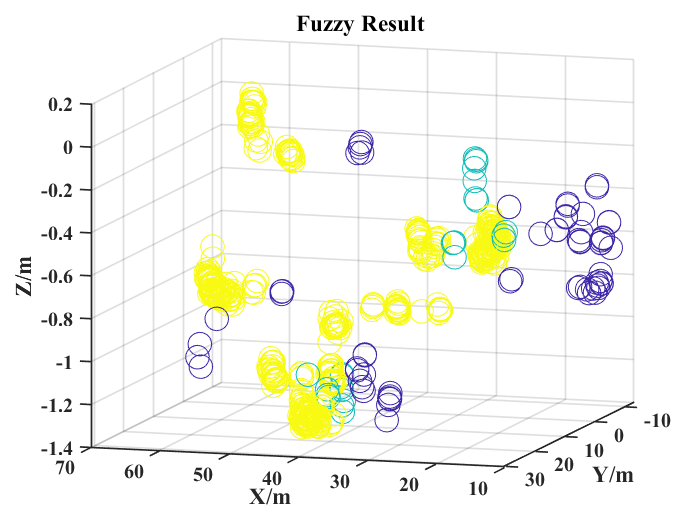}}

\caption{DBSCAN and fuzzy classification results on the candidate boxes. Different colors represent different categories. After Fuzzy clustering, multitudinous categories of DBSCAN are compressed into 3 categories.}
\label{fig7}
\end{figure}

\subsection{Experimental Details}
To verify the effectiveness of this approach, we selected popular 3D object detection models, including PointPillars \cite{lang2019pointpillars}, SECOND \cite{yan2018second}, \textcolor{black}{etc}, used as baselines from the OpenPCDet framework. In this baseline training process, ADAM was selected as the optimizer, and cosine annealing was implemented as the optimization strategy. In the DBSCAN clustering stage, neighborhood parameters were set in DBSCAN for the candidate bounding boxes generated by the model, selecting a neighborhood of 0.3, with four samples for the MinPts threshold. Membership functions for volume, density, and classification were then established using the fuzzy system. \mbox{Table \ref{table_3}} provided membership function parameters for each fuzzy set variable, utilizing the membership function shown in \mbox{Fig. \ref{fig6}}. This information was then used to divide the fuzzy set and finally perform fuzzy logic reasoning, applying the fuzzy rules from \mbox{Table \ref{table_2}} to classify candidate boxes. To further analyze the role of the clustering algorithm and fuzzy classification system in NMS, the centers of candidate bounding boxes were visualized prior to algorithm execution, as shown in \mbox{Fig. \ref{fig7}}. \mbox{Fig. \ref{fig7}}(a) showed the results of DBSCAN clustering, where each color represented a different cluster. As shown, the bounding boxes with different densities were grouped into a single class. Boxes were also divided into clusters based on the spatial distance, with boxes in the same cluster exhibiting the same density. \mbox{Fig. \ref{fig7}}(b) indicated the results of fuzzy classification fell into three distinct categories: yellow, cyan, and purple, corresponding to LVHD, SVHD, and LD, respectively. These clusters were then further grouped into new clusters, thus achieving the goal of classifying boxes by volume and density as part of candidate box screening. After fuzzy classification, the three box categories were assigned different thresholds, as shown in \mbox{Table \ref{table_4}}. 

\begin{table*}[]
	\centering
    % \fontsize{2.0pt}{\baselineskip}\selectfont%设置字体大小
	\caption{\textcolor{black}{Comparative experiments on the KITTI 3D object detection test benchmark. In order to ensure fairness, all weight models we use for testing are downloaded from the official model zoo of each baseline. ``Mod." means moderate.}}
    \label{table_5} 
	\renewcommand\arraystretch{1}%行高
	% \resizebox{\linewidth}{!}{
\begin{tabularx}{\textwidth}{r|c|XXXXXXXXXXXX}
\specialrule{0.8pt}{0pt}{0pt}%第一行线宽
\multirow{2}{*}{Method}& \multirow{2}{*}{Publication} & \multicolumn{3}{c}{Average} & \multicolumn{3}{c}{Car}    & \multicolumn{3}{c}{Pedestrian}             & \multicolumn{3}{c}{Cyclist}                \\
                 &       & Easy            & Mod.     & Hard    & Easy            & Mod.             & Hard            & Easy            & Mod.             & Hard        & Easy            & Mod.     & Hard     \\ \hline
PointPillars\cite{lang2019pointpillars} &CVPR2019  &64.12 		&53.28 		&48.25                 &82.31 		&72.98 		&66.73 		&40.65 		&32.74 		&30.19 		&69.39 		&54.12 	&47.82 		  \\
+Fuzzy-NMS    &             &64.44 		&53.99 		&48.96 		&82.45 		&73.13 		&66.90 		&41.81 		&34.16 		&31.63 		&69.07 		&54.68 		&48.34          \\
\rowcolor{mygreen} Delta     &              &\textbf{+0.32} 		&\textbf{+0.71} 		&\textbf{+0.71} 		&+0.14 		&+0.15 		&+0.17 		&\textbf{+1.16} 		&\textbf{+1.42} 		&\textbf{+1.44} 		&-0.32 		&\textbf{+0.56} 		&\textbf{+0.52}  \\ \hline
PV-RCNN\cite{shi2020pv}    &CVPR2020         &71.00 		&59.91 		&54.85 		&87.11 		&78.62 		&73.97 		&47.99 		&39.55 		&36.49 		&77.91 		&61.56 		&54.09          \\
+Fuzzy-NMS        &          &71.07 		&60.26 		&55.51 		&87.30 		&78.71 		&74.03 		&48.38 		&40.01 		&37.31 		&77.53 		&62.05 		&55.18          \\
\rowcolor{mygreen} Delta      &            &0.07 		&\textbf{+0.35} 		&\textbf{+0.66} 		&+0.19 		&+0.09 		&+0.06 		&\textbf{+0.39} 		&\textbf{+0.46} 		&\textbf{+0.82} 		&-0.38 		&\textbf{+0.49} 		&\textbf{+1.09}           \\  \hline
IA-SSD\cite{zhang2022not}   &CVPR2022       &68.44 		&58.43 		&53.99 		&87.09 		&78.86 		&73.88 		&45.59 		&37.44 		&34.98 		&72.64 		&58.98 		&53.12          \\
+Fuzzy-NMS     &             &70.49 		&59.72 		&54.48 		&87.10 		&78.93 		&72.20 		&48.15 		&39.95 		&36.77 		&76.21 		&60.28 		&54.48         \\
\rowcolor{mygreen} Delta    &            &\textbf{+2.05} 		&\textbf{+1.29} 		&\textbf{+0.49} 		&+0.01 		&+0.07 		&-1.68 		&\textbf{+2.56} 		&\textbf{+2.51} 		&\textbf{+1.79} 		&\textbf{+3.61} 		&\textbf{+1.30} 		&\textbf{+1.36} \\ \hline
GD-MAE\cite{yang2022gd}   &CVPR2023       &67.47 		&55.74 		&51.03 		&87.01 		&76.04 		&70.73 		&45.49 		&36.07 		&32.85 		&69.91 		&55.11 		&49.52          \\
+Fuzzy-NMS       &           &67.60 		&55.83 		&51.42 		&87.00 		&76.03 		&70.72 		&46.00 		&36.46 		&34.14 		&69.80 		&54.99 		&49.39          \\
\rowcolor{mygreen} Delta     &           &+0.13 		&+0.09 		&\textbf{+0.39} 		&-0.01 		&-0.01 		&-0.01 		&\textbf{+0.51} 		&\textbf{+0.39} 		&\textbf{+1.29} 		&-0.11 		&-0.12 		&-0.13\\ \hline
BiProDet\cite{zhang2023bidirectional}  &ICLR2023        &73.24 		&63.74 		&58.75 		&87.68 		&81.58 		&76.82 		&52.77 		&45.42 		&41.77 		&79.27 		&64.21 		&57.65          \\
+Fuzzy-NMS     &             &73.81 		&63.83 		&58.80 		&88.10 		&81.77 		&76.90 		&52.91 		&45.71 		&42.06 		&80.42 		&64.01 		&57.43          \\
\rowcolor{mygreen} Delta      &          &\textbf{+0.57} 		&+0.09 		&+0.05 		&\textbf{+0.42} 		&\textbf{+0.19} 		&+0.08 		&\textbf{+0.14} 		&\textbf{+0.29} 		&\textbf{+0.29} 		&\textbf{+1.15} 		&-0.20 		&-0.22\\
\bottomrule[0.8pt]
\end{tabularx}%}
\end{table*}

\begin{table*}[]%*表示为跨栏表格，*去掉为单栏表格。
    \setlength\tabcolsep{3pt}
	\centering
	% \fontsize{2.0pt}{\baselineskip}\selectfont%设置字体大小
	\caption{\textcolor{black}{Comparative experiments on the KITTI BEV object detection test benchmark. In order to ensure fairness, all weight models we use for testing are downloaded from the official model zoo of each baseline. ``Mod." means moderate.}}
    \label{table_6}
	\renewcommand\arraystretch{1}%行宽
	% \resizebox{\linewidth}{!}{
\begin{tabularx}{\textwidth}{r|c|XXXXXXXXXXXX}%c的个数代表列的个数
\specialrule{0.8pt}{0pt}{0pt}%第一行线宽
\multirow{2}{*}{Method}& \multirow{2}{*}{Publication} & \multicolumn{3}{c}{Average}& \multicolumn{3}{c}{Car}                   & \multicolumn{3}{c}{Pedestrian}            & \multicolumn{3}{c}{Cyclist}               \\
             &           & \hspace{0.7em}Easy            & \hspace{0.7em}Mod.             & \hspace{0.6em}Hard            & \hspace{0.7em}Easy            & \hspace{0.7em}Mod.             &\hspace{0.3em} Hard            &\hspace{0.4em} Easy            & \hspace{0.7em}Mod.             & \hspace{0.6em}Hard            & \hspace{0.7em}Easy            & \hspace{0.6em}Mod.             & \hspace{0.6em}Hard 
                        \\ \hline
PointPillars\cite{lang2019pointpillars}     &CVPR2019     &\hspace{0.5em}70.33 		&\hspace{0.5em}62.06 		&\hspace{0.5em}57.43 		& \hspace{0.5em}90.10 		&\hspace{0.5em}86.60 		&\hspace{0.5em}81.72 		&\hspace{0.5em}45.47 		&\hspace{0.5em}37.88 		&\hspace{0.5em}35.46  	&\hspace{0.5em}75.43		&\hspace{0.5em}61.71 		&\hspace{0.5em}55.12          \\
+Fuzzy-NMS        &        &\hspace{0.5em}70.94 		&\hspace{0.5em}62.91 		&\hspace{0.5em}58.31 		& \hspace{0.5em}90.27 		&\hspace{0.5em}86.79 		&\hspace{0.5em}81.91 		&\hspace{0.5em}47.01 		&\hspace{0.5em}39.74 		&\hspace{0.5em}37.30 		&\hspace{0.5em}75.53 		&\hspace{0.5em}62.20 		&\hspace{0.5em}55.73         \\
\rowcolor{mygreen} Delta    &             & \hspace{0.5em}\textbf{+0.61} 		& \hspace{0.5em}\textbf{+0.85} 		& \hspace{0.5em}\textbf{+0.88} 		&\hspace{0.5em}+0.17 		&\hspace{0.5em}+0.19 		&\hspace{0.5em}+0.19 		&\hspace{0.5em}\textbf{+1.54} 		&\hspace{0.5em}\textbf{+1.86} 		&\hspace{0.5em}\textbf{+1.84} 		&\hspace{0.5em}+0.10 		&\hspace{0.5em}\textbf{+0.49} 		&\hspace{0.5em}\textbf{+0.61}   \\  \hline
PV-RCNN\cite{shi2020pv}   &CVPR2020            &\hspace{0.5em}75.17 		&\hspace{0.5em}65.75 		&\hspace{0.5em}61.67 		&\hspace{0.5em}91.64 		&\hspace{0.5em}87.69 		&\hspace{0.5em}84.60 		&\hspace{0.5em}53.22 		&\hspace{0.5em}44.93 		&\hspace{0.5em}42.50 		&\hspace{0.5em}80.66 		&\hspace{0.5em}64.62 		&\hspace{0.5em}57.92          \\
+Fuzzy-NMS      &            &\hspace{0.5em}75.50 		&\hspace{0.5em}66.40 		&\hspace{0.5em}61.82 		&\hspace{0.5em}91.76 		&\hspace{0.5em}87.71 		&\hspace{0.5em}83.17 		&\hspace{0.5em}53.57 		&\hspace{0.5em}45.37 		&\hspace{0.5em}42.72 		&\hspace{0.5em}81.16 		&\hspace{0.5em}66.12 		&\hspace{0.5em}59.56          \\
\rowcolor{mygreen} Delta    &               &\hspace{0.5em}\textbf{+0.33} 		&\hspace{0.5em}\textbf{+0.65} 		&\hspace{0.5em}+0.15 		&\hspace{0.5em}+0.12 		&\hspace{0.5em}+0.02 		&\hspace{0.5em}-1.43 		&\hspace{0.5em}\textbf{+0.35} 		&\hspace{0.5em}\textbf{+0.44} 		&\hspace{0.5em}+0.22 		&\hspace{0.5em}\textbf{+0.50} 		&\hspace{0.5em}\textbf{+1.50} 		&\hspace{0.5em}\textbf{+1.64}            \\  \hline
IA-SSD\cite{zhang2022not}   &CVPR2022       &\hspace{0.5em}74.46 		&\hspace{0.5em}66.31 		&\hspace{0.5em}61.24 		&\hspace{0.5em}92.33		&\hspace{0.5em}88.66 		&\hspace{0.5em}83.74 		&\hspace{0.5em}51.49 		&\hspace{0.5em}43.56 		&\hspace{0.5em}40.37 		&\hspace{0.5em}79.55 		&\hspace{0.5em}66.72 		&\hspace{0.5em}59.61          \\
+Fuzzy-NMS      &            &\hspace{0.5em}76.22 		&\hspace{0.5em}67.47 		&\hspace{0.5em}62.79 		&\hspace{0.5em}92.47 		&\hspace{0.5em}89.01 		&\hspace{0.5em}84.00 		&\hspace{0.5em}53.48 		&\hspace{0.5em}45.68 		&\hspace{0.5em}43.14 		&\hspace{0.5em}82.70 		&\hspace{0.5em}67.72 		&\hspace{0.5em}61.23          \\
\rowcolor{mygreen} Delta   &             &\hspace{0.5em}\textbf{+1.76} 		&\hspace{0.5em}\textbf{+1.16} 		&\hspace{0.5em}\textbf{+1.55} 		&\hspace{0.5em}+0.14 		&\hspace{0.5em}\textbf{+0.35} 		&\hspace{0.5em}+0.26 		&\hspace{0.5em}\textbf{+1.99} 		&\hspace{0.5em}\textbf{+2.12} 		&\hspace{0.5em}\textbf{+2.77} 		&\hspace{0.5em}\textbf{+3.15}		&\hspace{0.5em}\textbf{+1.00} 		&\hspace{0.5em}\textbf{+1.62} \\ \hline
GD-MAE\cite{yang2022gd}     &CVPR2023     &\hspace{0.5em}73.58 		&\hspace{0.5em}63.94 		&\hspace{0.5em}58.67 		&\hspace{0.5em}91.81 		&\hspace{0.5em}88.39 		&\hspace{0.5em}83.23 		&\hspace{0.5em}49.69 		&\hspace{0.5em}40.87 		&\hspace{0.5em}37.64 		&\hspace{0.5em}79.23 		&\hspace{0.5em}62.57 		&\hspace{0.5em}55.14          \\
+Fuzzy-NMS       &           &\hspace{0.5em}74.02 		&\hspace{0.5em}64.05 		&\hspace{0.5em}59.12 		&\hspace{0.5em}91.79 		&\hspace{0.5em}88.38 		&\hspace{0.5em}83.21 		&\hspace{0.5em}51.18 		&\hspace{0.5em}41.32 		&\hspace{0.5em}39.10 		&\hspace{0.5em}79.08 		&\hspace{0.5em}62.46 		&\hspace{0.5em}55.05          \\
\rowcolor{mygreen} Delta      &          &\hspace{0.5em}\textbf{+0.44} 		&\hspace{0.5em}+0.11 		&\hspace{0.5em}\textbf{+0.45} 		&\hspace{0.5em}-0.02 		&\hspace{0.5em}-0.01 		&\hspace{0.5em}-0.02 		&\hspace{0.5em}\textbf{+1.49} 		&\hspace{0.5em}\textbf{+0.45} 		&\hspace{0.5em}\textbf{+1.46} 		&\hspace{0.5em}-0.15 		&\hspace{0.5em}-0.11 		&\hspace{0.5em}-0.09\\ \hline
BiProDet\cite{zhang2023bidirectional}   &ICLR2023       &\hspace{0.5em}77.86 		&\hspace{0.5em}68.90 		&\hspace{0.5em}64.81 		&\hspace{0.5em}92.16 		&\hspace{0.5em}89.05 		&\hspace{0.5em}86.05 		&\hspace{0.5em}57.99 		&\hspace{0.5em}49.32 		&\hspace{0.5em}46.89 		&\hspace{0.5em}83.42 		&\hspace{0.5em}68.32 		&\hspace{0.5em}61.49          \\
+Fuzzy-NMS     &             &\hspace{0.5em}78.73 		&\hspace{0.5em}70.11 		&\hspace{0.5em}64.41 		&\hspace{0.5em}92.21 		&\hspace{0.5em}89.02 		&\hspace{0.5em}84.10 		&\hspace{0.5em}58.41 		&\hspace{0.5em}50.97 		&\hspace{0.5em}47.31 		&\hspace{0.5em}85.57 		&\hspace{0.5em}70.34 		&\hspace{0.5em}61.81          \\
\rowcolor{mygreen} Delta   &             &\hspace{0.5em}\textbf{+0.87} 		&\hspace{0.5em}\textbf{+1.21} 		&\hspace{0.5em}-0.40 		&\hspace{0.5em}+0.05 		&\hspace{0.5em}-0.03 		&\hspace{0.5em}-1.95 		&\hspace{0.5em}\textbf{+0.42} 		&\hspace{0.5em}\textbf{+1.65} 		&\hspace{0.5em}\textbf{+0.42} 		&\hspace{0.5em}\textbf{+2.15} 		&\hspace{0.5em}\textbf{+2.02} 		&\hspace{0.5em}+0.32\\
\bottomrule[0.8pt]
\end{tabularx}%}
\end{table*}

\subsection{Experiment Results and Analysis}
For a fair comparison, it was proved that the improvement of the algorithm was caused by our proposed Fuzzy-NMS module. We conducted experiments on the \textcolor{black}{KITTI 3D object detection test and validation sets and large-scale Waymo 3D object detection dataset. The trained model we used was provided by the official model zoo}. We just replaced the original NMS module with our proposed Fuzzy-NMS without training again. This reduced the impact of uncertain factors such as network training. \textcolor{black}{Therefore, baseline detection results in the KITTI test set were benchmarked against the model submission results provided by the official model zoo.}

\noindent
\textcolor{black}{\textbf{KITTI test set.} First, we validated the performance of our Fuzzy-NMS on five different baselines on the KITTI test set. These baselines were the five representative ones we found from 2019 to 2023 after careful selection, namely PointPillars(CVPR2019)\cite{lang2019pointpillars}, PV-RCNN(CVPR2020)\cite{shi2020pv}, IA-SSD(CVPR2022)\cite{zhang2022not}, GD-MAE(CVPR2023)\cite{yang2022gd} and BiProDet(ICLR2023)\cite{zhang2023bidirectional}. The weight models we used for testing were provided by the official of each baseline, which could avoid various effects caused by repeated training, and we just replaced the original NMS module with the Fuzzy-NMS module. The results of the comparison experiment between the original performance of each baseline and the performance after adding Fuzzy-NMS are shown in \mbox{Table \ref{table_5}} and \mbox{Table \ref{table_6}}, where \mbox{Table \ref{table_5}} shows the results of 3D detection indicators and \mbox{Table \ref{table_6}} shows the results of BEV. As can be seen from the two tables, the addition of the Fuzzy-NMS module has brought improvements to the indicators of each baseline on 3D and BEV. For example, our method exhibited average 3D performance improvements of 1.29$\%$ and 0.71$\%$, and BEV performance improvements of 1.16$\%$ and 0.85$\%$ at a moderate level for three-category averages on IA-SSD \cite{zhang2022not} and PointPillars \cite{lang2019pointpillars}.}

\noindent
\textcolor{black}{\textbf{KITTI validation set.} Second, in order to prove the applicability of our method on more baselines, we conducted comparative experiments on nine different baselines on the KITTI validation set. These nine baselines included different classic works from 2018 to 2023. TABLE VII and TABLE VIII show the comparative results on the 3D and BEV benchmarks, respectively. Similar to the experiments on the test set, the results on the validation set were also satisfactory to us. Whether it is on the old baselines four or five years ago or on the advanced baselines in the past two years, the improvement effect brought by Fuzzy-NMS is very significant. And this improvement is mainly reflected in small objects such as pedestrians and cyclists. The 3D and BEV benchmarks of most baselines at the moderate level of Pedestrian and Cyclist can increase by 0.50$\%$ or more, of which some even increases by 2$\%$ or more than 3$\%$, like GD-MAE \cite{yang2022gd}, IA-SSD \cite{zhang2022not}, PointPillars \cite{lang2019pointpillars}. It can be seen  that our Fuzzy-NMS module has been proven to perform well on the KITTI dataset, and the baseline selection with a time span of up to five years also proves that our method is universal.}

\noindent
\textcolor{black}{\textbf{Waymo validation set.} In addition, to prove that the Fuzzy-NMS module performs well on larger datasets, we also carried out comparative experiments on the validation set of the Waymo Open Dataset. TABLE IX shows the experimental results on Waymo. Similar to the results on the KITTI validation set and test set, the addition of Fuzzy-NMS allowed the model to obtain higher accuracy in the detection of the Pedestrian category, and the improvement is more obvious. The increase in the benchmark on the SECOND [ 8] baseline can even reach more than 5$\%$. So it can be seen that the generality of our method on different data sets can be guaranteed.}
%~\\\\

\begin{table*}[]

    \centering
    % \fontsize{5.0pt}{\baselineskip}\selectfont%设置字体大小
    \caption{\textcolor{black}{Comparative experiments on the KITTI 3d object detection validation benchmark. All baselines are from the trained models of the OpenPCDet framework. ``Mod." means moderate.}}
    \label{table_7} 
	\renewcommand\arraystretch{1}%行高
	% \resizebox{\linewidth}{!}{
\begin{tabularx}{\textwidth}{r|c|XXXXXXXXXXXX}
\specialrule{0.8pt}{0pt}{0pt}%第一行线宽
\multirow{2}{*}{Method} & \multirow{2}{*}{Publication}  &\multicolumn{3}{c}{Average} & \multicolumn{3}{c}{Car}    & \multicolumn{3}{c}{Pedestrian}             & \multicolumn{3}{c}{Cyclist}                \\
                &        & Easy            & Mod.     & Hard    & Easy            & Mod.             & Hard            & Easy            & Mod.             & Hard        & Easy            & Mod.     & Hard     \\ \hline
SECOND\cite{yan2018second}       &Sensors2018           &76.48 		&66.50 		&62.52 		&90.55 		&81.61 		&78.61 		&55.94 		&51.14 		&46.17 		&82.96 		&66.74 		&62.78          \\
+Fuzzy-NMS    &            &77.22 		&67.33 		&63.33 		&90.91 		&82.01 		&78.93 		&57.32 		&52.64 		&47.62 		&83.43 		&67.34 		&63.44          \\
\rowcolor{mygreen} Delta          &         &\textbf{+0.74} 		&\textbf{+0.83} 		&\textbf{+0.81} 		&\textbf{+0.36} 		&\textbf{+0.40} 		&+0.32 		&\textbf{+1.38} 		&\textbf{+1.50} 		&\textbf{+1.45} 		&\textbf{+0.47} 		&\textbf{+0.60} 		&\textbf{+0.66}  \\ \hline
PointPillars\cite{lang2019pointpillars}    &CVPR2019        &75.54 		&64.21 		&60.29 		&87.75 		&78.40 		&75.18 		&57.30 		&51.41 	&46.87 		&81.57 		&62.81 		&58.83         \\
+Fuzzy-NMS    &            &76.42 		&65.19 		&61.34 		&88.09 		&78.77 		&75.45 		&59.45 		&53.62 		&49.06 		&81.73 		&63.17 		&59.50          \\
\rowcolor{mygreen} Delta     &              &\textbf{+0.88} 		&\textbf{+0.98} 		&\textbf{+1.04} 		&\textbf{+0.34} 		&\textbf{+0.37} 		&+0.27 		&\textbf{+2.15} 		&\textbf{+2.21} 		&\textbf{+2.19} 		&+0.16 		&\textbf{+0.36} 		&\textbf{+0.67}  \\ \hline
PV-RCNN\cite{shi2020pv}      &CVPR2020        &81.30 		&69.77 		&66.13 		&92.11 		&84.39 		&82.50 		&62.70 		&54.51 		&49.86 		&89.10 		&70.40 		&66.02          \\
+Fuzzy-NMS     &             &81.79 		&70.32 		&66.72 		&92.12 		&84.47 		&82.62 		&63.62 		&55.79 		&51.11 		&89.62 		&70.70 		&66.43          \\
\rowcolor{mygreen} Delta     &             &\textbf{+0.48} 		&\textbf{+0.55} 		&\textbf{+0.59} 		&+0.01 		&+0.08 		&+0.12 		&\textbf{+0.92} 		&\textbf{+1.28} 		&\textbf{+1.25} 		&\textbf{+0.52} 		&+0.30 		&\textbf{+0.41}           \\ \hline
Part-A2-Anchor\cite{shi2020points}    &TPAMI2020      &83.13 		&70.91 		&67.85 		&92.15 		&82.91 		&82.00 		&66.89 		&59.68 		&54.62 		&90.34 		&70.14 		&66.93          \\
+Fuzzy-NMS      &            &83.07 		&71.57 		&68.42 		&92.39 		&83.18 		&82.25 		&67.35 		&60.28 		&55.14 		&89.47 		&71.25 		&67.86       \\
\rowcolor{mygreen} Delta    &             &-0.06 		&\textbf{+0.66} 		&\textbf{+0.57} 		&+0.24 		&+0.27 		&+0.25 		&\textbf{+0.46} 		&\textbf{+0.60} 		&\textbf{+0.52} 		&-0.87 		&\textbf{+1.11} 		&\textbf{+0.93} \\ \hline
Part-A2-Free\cite{shi2020points}      &TPAMI2020    &85.28 		&74.00 		&69.61 		&91.66 		&80.29 		&78.08 		&72.31 		&66.37 		&60.07 		&91.86 		&75.33 		&70.68          \\
+Fuzzy-NMS      &             &86.21 		&74.56 		&70.64 		&92.23 		&80.53 		&78.31 		&73.09 		&66.69 		&61.14 		&93.31 		&76.45 		&72.47         \\
\rowcolor{mygreen} Delta   &             &\textbf{+0.93} 		&\textbf{+0.56} 		&\textbf{+1.03} 		&\textbf{+0.57} 		&+0.24 		&+0.23 		&\textbf{+0.78} 		&+0.32 		&\textbf{+1.07} 		&\textbf{+1.45} 		&\textbf{+1.12} 		&\textbf{+1.79} \\ \hline
IA-SSD\cite{zhang2022not}   &CVPR2022      &81.34 		&70.47 		&66.38 		&91.71 		&83.27 		&80.26 		&61.51 		&56.42 		&51.30 		&90.79 		&71.73 		&67.58          \\
+Fuzzy-NMS        &             &82.11 		&71.52 		&67.31 		&91.94 		&83.54 		&80.50 		&63.56 		&59.22 		&53.82 		&90.82 		&71.80 		&67.62       \\
\rowcolor{mygreen} Delta    &            &\textbf{+0.77} 		&\textbf{+1.05} 		&\textbf{+0.93} 		&+0.23 		&+0.27 		&+0.24 		&\textbf{+2.05} 		&\textbf{+2.80} 		&\textbf{+2.52} 		&+0.03 		&+0.07 		&+0.04 \\\hline
M3DETR\cite{guan2022m3detr}    &WACV2022      &81.52 		&70.20 		&66.01 		&91.46 		&84.37 		&81.61 		&61.91 		&53.57 		&48.49 		&91.18 		&72.67 		&67.93          \\
+Fuzzy-NMS        &          &82.01 		&70.61 		&66.20 		&91.69 		&84.66 		&81.87 		&63.16 		&54.59 		&48.88 		&91.17 		&72.59 		&67.85          \\
\rowcolor{mygreen} Delta   &             &\textbf{+0.49} 		&\textbf{+0.41} 		&+0.19 		&+0.23 		&+0.29 		&+0.26 		&\textbf{+1.25} 		&\textbf{+1.02} 		&\textbf{+0.39} 		&-0.01 		&-0.08		&-0.08\\ \hline
GD-MAE\cite{yang2022gd}  &CVPR2023        &77.21 		&65.84 		&62.13 		&91.49 		&82.01 		&79.08 		&52.05 		&48.40 		&44.65 		&88.10 		&67.12 		&62.67          \\
+Fuzzy-NMS      &            &77.77 		&66.23 		&62.53 		&91.48 		&82.00 		&79.07 		&54.01 		&49.70 		&45.88 		&87.82 		&66.98 		&62.63          \\
\rowcolor{mygreen} Delta  &              &\textbf{+0.56} 		&\textbf{+0.38} 		&\textbf{+0.40} 		&-0.01 		&-0.01 		&-0.01 		&\textbf{+1.96} 		&\textbf{+1.30} 		&\textbf{+1.23} 		&-0.28 		&-0.14 		&-0.04\\ \hline
BiProDet\cite{zhang2023bidirectional}  &ICLR2023        &86.79 		&76.88 		&72.74 		&92.54 		&86.04 		&83.42 		&72.37 		&67.45 		&62.46 		&95.46 		&77.15 		&72.34          \\
+Fuzzy-NMS     &             &87.23 		&77.36 		&73.22 		&92.53 		&86.04 		&83.41 		&73.70 		&68.89 		&63.90 		&95.56 		&77.16 		&72.36          \\
\rowcolor{mygreen} Delta     &           &\textbf{+0.44} 		&\textbf{+0.48} 		&\textbf{+0.48} 		&-0.01 		&+0.00 		&-0.01 		&\textbf{+1.33} 		&\textbf{+1.44} 		&\textbf{+1.44} 		&+0.00 		&+0.01 		&+0.02\\\bottomrule[0.8pt]
\end{tabularx}%}
\end{table*}

\noindent
\textcolor{black}{\textbf{Overall analysis.} In conclusion, the Fuzzy-NMS module performs well on small objects such as pedestrians and cyclists, which can bring a lot of improvements compared to the original NMS module. However, it is undeniable that our module also has certain shortcomings, that is, the performance on large objects such as cars is not good, and many benchmarks have not been improved. Some are even lower than the original ones. After analysis, we consider this is because the Fuzzy-NMS module classifies objects with different volume densities and selects appropriate IoU and score thresholds for each category, and when a large object like a car is driving or parking, two objects rarely overlap each other, so it is difficult to improve the accuracy by adjusting the threshold of NMS for the car category. On the contrary, small objects like pedestrians or cyclists are more likely to overlap, and the pre-selected boxes of overlapping objects are too dense, and the original NMS module often misses small objects due to setting a constant threshold. Therefore, when Fuzzy-NMS is added, the bounding box that was originally wrongly suppressed can be correctly screened out through reasonable adjustment of the NMS threshold. In real life, pedestrians and cyclists are often unpredictable in their actions and often cause many traffic accidents, so the successful detection of pedestrians and cyclists is very important in the field of automatic driving. So although our module actually doesn’t achieve good results in every category, their improvement in dense small objects is also worthy of attention.}

%fig8
\begin{figure}[!t]
	\vspace{-5mm}
	\centering
	\includegraphics[width=1\linewidth,height=9cm]{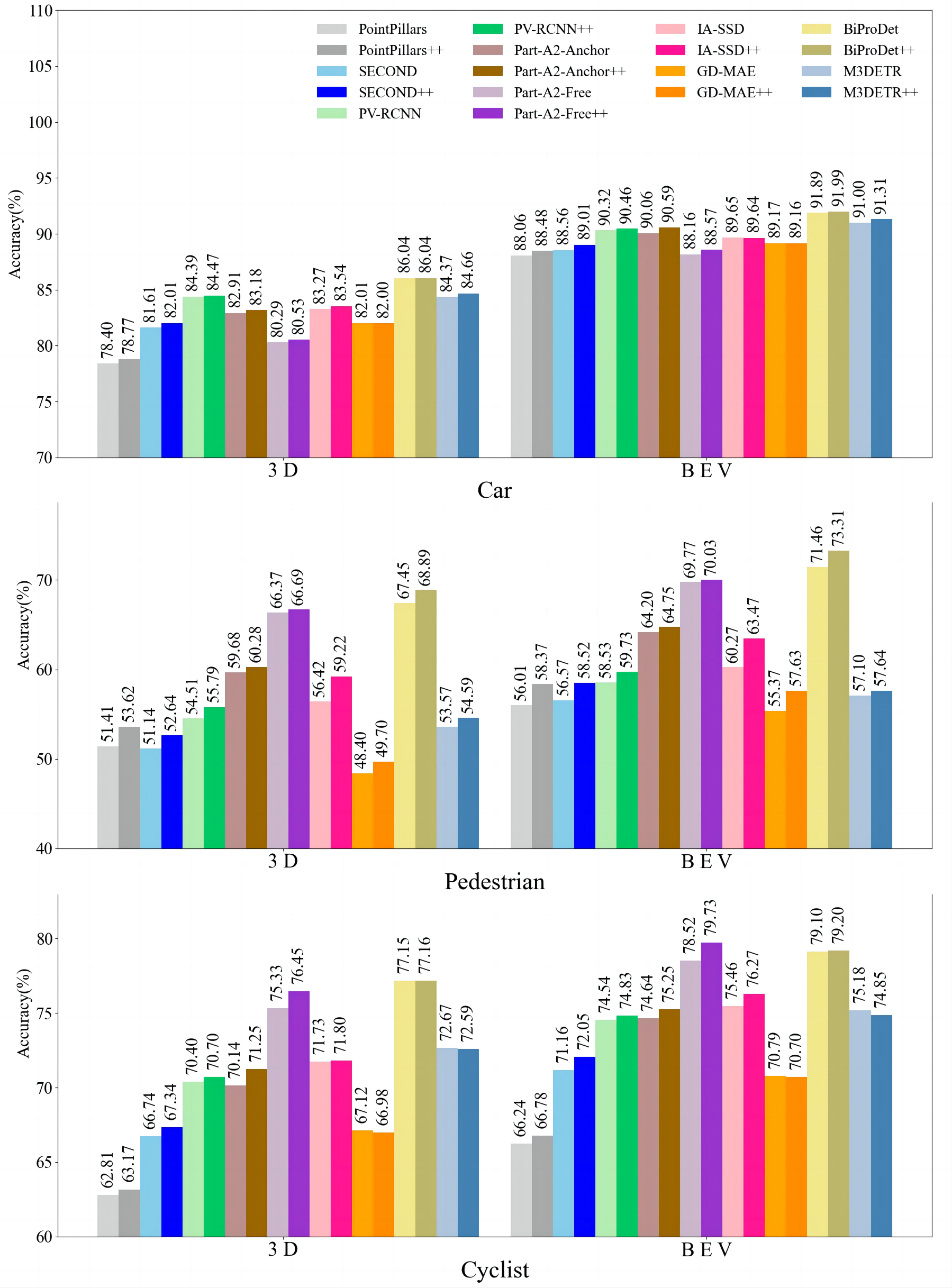}
	\caption{\textcolor{black}{Contrastive visualization results of five models at the moderate level. In the legend, ‘++’ represents the baseline model with the Fuzzy-NMS, and light colors represent the results of the original models, while dark colors represent the corresponding models with the Fuzzy-NMS module.
		}}
	\label{fig8}
\end{figure}

\begin{table*}[]%*表示为跨栏表格，*去掉为单栏表格。
	\centering
	% \fontsize{5.0pt}{\baselineskip}\selectfont%设置字体大小
	\caption{\textcolor{black}{Comparative experiments on the KITTI BEV object detection validation benchmark. All baselines are from the trained models of the OpenPCDet framework. ``Mod." means moderate.}}
    \label{table_8}
	\renewcommand\arraystretch{1}%行高
	% \resizebox{\linewidth}{!}{
\begin{tabularx}{\textwidth}{r|c|XXXXXXXXXXXX}%c的个数代表列的个数
\specialrule{0.8pt}{0pt}{0pt}%第一行线宽
\multirow{2}{*}{Method} & \multirow{2}{*}{Publication}  & \multicolumn{3}{c}{Average}& \multicolumn{3}{c}{Car}                   & \multicolumn{3}{c}{Pedestrian}            & \multicolumn{3}{c}{Cyclist}              \\
                  &      & Easy            & Mod.             & Hard            & Easy            & Mod.             & Hard            & Easy            & Mod.             & Hard            & Easy            & Mod.             & Hard 
                        \\ \hline
SECOND\cite{yan2018second}       &Sensors2018         &80.40 		&72.10 		&68.89 		&92.42 		&88.56 		&87.65 		&60.73 		&56.57 		&52.14 		&88.05 		&71.16 		&66.89          \\
+Fuzzy-NMS    &              &80.94 		&73.19 		&69.93 		&92.84 		&89.01 		&88.00 		&62.54 		&58.52 		&54.11 		&87.44 		&72.05 		&67.69         \\
\rowcolor{mygreen} Delta       &            &\textbf{+0.54 }		&\textbf{+1.09 }		&\textbf{+1.04 }		&\textbf{+0.42 }		&\textbf{+0.45} 		&\textbf{+0.35} 		&\textbf{+1.81} 		&\textbf{+1.95} 		&\textbf{+1.97} 		&-0.61		&\textbf{+0.89} 		&\textbf{+0.80 } \\ \hline
PointPillars\cite{lang2019pointpillars}   &CVPR2019        &79.63 		&70.10 		&66.98 		& 92.04 		&88.06 		&86.66 		&61.60 		&56.01 		&52.05  	&85.26		&66.24 		&62.22          \\
+Fuzzy-NMS         &         &80.81 		&71.21 		&67.72 		& 92.45 		&88.48 		&85.80 		&64.11 		&58.37 		&54.39 		&85.87 		&66.78 		&62.98       \\
\rowcolor{mygreen} Delta      &            & \textbf{+1.18} 		& \textbf{+1.11} 		& \textbf{+0.74} 		& \textbf{+0.41} 		&\textbf{+0.42} 		&-0.86 		&\textbf{+2.51} 		&\textbf{+2.36} 		&\textbf{+2.34} 		&\textbf{+0.61} 		&\textbf{+0.54} 		&\textbf{+0.76}   \\ \hline
PV-RCNN\cite{shi2020pv}    &CVPR2020            &84.15 		&74.46 		&70.92 		&93.03 		&90.32 		&88.53 		&65.93 		&58.53 		&54.13 		&93.48 		&74.54 		&70.10          \\
+Fuzzy-NMS     &             &84.35 		&75.01 		&71.76 		&93.03 		&90.46 		&88.69 		&67.28 		&59.73 		&55.43 		&92.74 		&74.83 		&71.15          \\
\rowcolor{mygreen} Delta    &                  &\textbf{+0.20} 		&\textbf{+0.55} 		&\textbf{+0.84} 		&0.00 		&+0.14 		&+0.16 		&\textbf{+1.35} 		&\textbf{+1.20} 		&\textbf{+1.30} 		&-0.74 		&+0.29 		&\textbf{+1.05}            \\ \hline
Part-A2-Anchor\cite{shi2020points}    &TPAMI2020      &85.13 		&76.30 		&72.74 		&92.90 		&90.06 		&88.35 		&70.53 		&64.20 		&59.25 		&91.96 		&74.64 		&70.63          \\
+Fuzzy-NMS       &              &85.58 		&76.86 		&73.48 		&93.32 		&90.59 		&88.66 		&71.02 		&64.75 		&59.75 		&92.39 		&75.25 		&72.02       \\
\rowcolor{mygreen} Delta    &                &\textbf{+0.45} 		&\textbf{+0.56} 		&\textbf{+0.74} 		&\textbf{+0.42} 		&\textbf{+0.53} 		&+0.31 		&\textbf{+0.49} 		&\textbf{+0.55} 		&\textbf{+0.50} 		&\textbf{+0.43} 		&\textbf{+0.61} 		&\textbf{+1.39}  \\ \hline
Part-A2-Free\cite{shi2020points}     &TPAMI2020      &87.19 		&78.82 		&74.85 		&92.85 		&88.16 		&86.17 		&75.44 		&69.77 		&64.46 		&93.27 		&78.52 		&73.91          \\
+Fuzzy-NMS          &         &88.19 		&79.44 		&75.68 		&93.57 		&88.57 		&86.65 		&76.31 		&70.03 		&64.62 		&94.70 		&79.73 		&75.76         \\
\rowcolor{mygreen} Delta     &            & \textbf{+1.00 }		& \textbf{+0.62 }		& \textbf{+0.83 }		& \textbf{+0.72 }		&\textbf{+0.41} 		&\textbf{+0.48} 		&\textbf{+0.87} 		&+0.26 		&+0.16 		&\textbf{+1.43} 		&\textbf{+1.21} 		&\textbf{+1.85 } \\ \hline
IA-SSD\cite{zhang2022not}     &CVPR2022      &84.09 		&75.13 		&71.66 		&93.28 		&89.65 		&88.64 		&66.19 		&60.27 		&55.25 		&92.80 		&75.46 		&71.09          \\
+Fuzzy-NMS       &           &84.98 		&76.46 		&72.94 		&93.26 		&89.64 		&88.64 		&68.93 		&63.47 		&58.33 		&92.76 		&76.27 		&71.85          \\
\rowcolor{mygreen} Delta      &           &\textbf{+0.89} 		&\textbf{+1.33} 		&\textbf{+1.28} 		&-0.02 		&-0.01 		&+0.00 		&\textbf{+2.74} 		&\textbf{+3.20} 		&\textbf{+3.08} 		&-0.04 		&\textbf{+0.81} 		&\textbf{+0.76} \\\hline
M3DETR\cite{guan2022m3detr}    &WACV2022       &84.00 		&74.43 		&70.29 		&94.88 		&91.00 		&88.38 		&64.31 		&57.10 		&52.13 		&92.80 		&75.18 		&70.37          \\
+Fuzzy-NMS    &              &84.49 		&74.60 		&70.51 		&95.14 		&91.31 		&88.66 		&65.53 		&57.64 		&52.61 		&92.79 		&74.85 		&70.25          \\
\rowcolor{mygreen} Delta       &          &\textbf{+0.49} 		&+0.17 		&+0.22 		&+0.26 		&\textbf{+0.31} 		&+0.28 		&\textbf{+1.22} 		&\textbf{+0.54} 		&\textbf{+0.48} 		&-0.01 		&-0.33		&-0.12\\ \hline
GD-MAE\cite{yang2022gd}     &CVPR2023      &82.56 		&71.78 		&68.26 		&95.11 		&89.17 		&88.30 		&60.19 		&55.37 		&51.41 		&92.39 		&70.79 		&65.06          \\
+Fuzzy-NMS       &           &83.00 		&72.50 		&69.41 		&95.10 		&89.16 		&88.28 		&61.65 		&57.63 		&53.77 		&92.24 		&70.70 		&66.18          \\
\rowcolor{mygreen} Delta     &            &\textbf{+0.44} 		&\textbf{+0.72} 		&\textbf{+1.15} 		&-0.01 		&-0.01 		&-0.02 		&\textbf{+1.46} 		&\textbf{+2.26} 		&\textbf{+2.36} 		&-0.15 		&-0.09 		&\textbf{+1.12}\\ \hline
BiProDet\cite{zhang2023bidirectional}    &ICLR2023       &89.15 		&80.82 		&76.72 		&93.61 		&91.89 		&89.49 		&76.37 		&71.46 		&66.19 		&97.48 		&79.10 		&74.48          \\
+Fuzzy-NMS    &              &89.76 		&81.50 		&77.39 		&93.77 		&91.99 		&89.57 		&78.03 		&73.31 		&68.02 		&97.47 		&79.20 		&74.57          \\
\rowcolor{mygreen} Delta    &             &\textbf{+0.61} 		&\textbf{+0.68} 		&\textbf{+0.67} 		&+0.16 		&+0.10 		&+0.08 		&\textbf{+1.66} 		&\textbf{+1.85} 		&\textbf{+1.83} 		&-0.01 		&+0.10 		&+0.09\\
\bottomrule[0.8pt]
\end{tabularx}%}
\end{table*}

\begin{table*}[]%*表示为跨栏表格，*去掉为单栏表格。
    \setlength\tabcolsep{1pt}
	\centering
	% \fontsize{2.0pt}{\baselineskip}\selectfont%设置字体大小
	\caption{\textcolor{black}{Comparative experiments on the Waymo object detection validation benchmark. All baselines are from the trained models of the OpenPCDet framework.}}
    \label{table_9}
	\renewcommand\arraystretch{0.95}%行高
	% \resizebox{\linewidth}{!}{
\begin{tabularx}{\textwidth}{r|XXXXXXXXXXXX}%c的个数代表列的个数

\specialrule{0.8pt}{0pt}{0pt}%第一行线宽

\multirow{2}{*}{Method}  & \multicolumn{4}{c}{Vehicle}                   & \multicolumn{4}{c}{Pedestrian}            & \multicolumn{4}{c}{Cyclist}               \\
                        & \hspace{0.4em}L1mAP            & L1mAPH     &\hspace{0.4em}L2mAP     &L2mAPH       &\hspace{0.4em}L1mAP    &L1mAPH     &\hspace{0.1em}L2mAP     &L2mAPH           &\hspace{0.2em}L1mAP     & L1mAPH     &\hspace{0.2em}L2mAP     &L2mAPH               
                        \\ \hline
PointPillars\cite{lang2019pointpillars}               &\hspace{0.7em}70.43 		&\hspace{0.7em}69.83 		&\hspace{0.7em}62.18 		&\hspace{0.7em}61.64 		&\hspace{0.7em}66.21 		&\hspace{0.7em}46.32 		&\hspace{0.7em}58.18 		&\hspace{0.7em}40.64 		&\hspace{0.7em}55.26 		&\hspace{0.7em}51.75 		&\hspace{0.7em}53.18 		&\hspace{0.7em}49.80          \\
+Fuzzy-NMS                  &\hspace{0.7em}70.15 		&\hspace{0.7em}69.53 		&\hspace{0.7em}62.05 		&\hspace{0.7em}61.50 		&\hspace{0.7em}67.61 		&\hspace{0.7em}46.87 		&\hspace{0.7em}60.31 		&\hspace{0.7em}41.74 		&\hspace{0.7em}57.78 		&\hspace{0.7em}54.02 		&\hspace{0.7em}56.19 		&\hspace{0.7em}52.53          \\
\rowcolor{mygreen} Delta                   &\hspace{0.7em}-0.28 		&\hspace{0.7em}-0.30 		&\hspace{0.7em}-0.13 		&\hspace{0.7em}-0.14 		&\hspace{0.7em}\textbf{+1.40} 		&\hspace{0.7em}\textbf{+0.55} 		&\hspace{0.7em}\textbf{+2.13} 		&\hspace{0.7em}\textbf{+1.10} 		&\hspace{0.7em}\textbf{+2.52} 		&\hspace{0.7em}\textbf{+2.27} 		&\hspace{0.7em}\textbf{+3.01} 		&\hspace{0.7em}\textbf{+2.73}               \\ \hline
SECOND\cite{yan2018second}                &\hspace{0.7em}69.07 		&\hspace{0.7em}68.45 		&\hspace{0.7em}60.86 		&\hspace{0.7em}60.29 		&\hspace{0.7em}59.97 		&\hspace{0.7em}49.50 		&\hspace{0.7em}53.26 		&\hspace{0.7em}43.89 		&\hspace{0.7em}53.21 		&\hspace{0.7em}51.82 		&\hspace{0.7em}51.86 		&\hspace{0.7em}50.49          \\
+Fuzzy-NMS                 &\hspace{0.7em}69.07 		&\hspace{0.7em}68.44 		&\hspace{0.7em}60.85 		&\hspace{0.7em}60.29 		&\hspace{0.7em}65.15 		&\hspace{0.7em}53.73 		&\hspace{0.7em}58.29 		&\hspace{0.7em}47.98 		&\hspace{0.7em}53.72 		&\hspace{0.7em}52.24 		&\hspace{0.7em}52.35 		&\hspace{0.7em}50.91          \\
\rowcolor{mygreen} Delta                  &\hspace{0.7em}+0.00		&\hspace{0.7em}-0.01		&\hspace{0.7em}-0.01		&\hspace{0.7em}+0.00		&\hspace{0.7em}\textbf{+5.18} 		&\hspace{0.7em}\textbf{+4.23} 		&\hspace{0.7em}\textbf{+5.03} 		&\hspace{0.7em}\textbf{+4.09} 		&\hspace{0.7em}\textbf{+0.51} 		&\hspace{0.7em}\textbf{+0.42}		&\hspace{0.7em}\textbf{+0.49} 		&\hspace{0.7em}\textbf{+0.42} \\ \hline
M3DETR\cite{guan2022m3detr}          &\hspace{0.7em}76.64 		&\hspace{0.7em}75.99 		&\hspace{0.7em}69.21 		& \hspace{0.7em}68.57 		&\hspace{0.7em}64.93 		&\hspace{0.7em}55.96 		&\hspace{0.7em}58.18 		&\hspace{0.7em}49.98 		&\hspace{0.7em}67.99  	&\hspace{0.7em}66.60		&\hspace{0.7em}65.98 		&\hspace{0.7em}64.64          \\
+Fuzzy-NMS                &\hspace{0.7em}76.58 		&\hspace{0.7em}75.92 		&\hspace{0.7em}69.24 		&\hspace{0.7em}68.60 		&\hspace{0.7em}67.10 		&\hspace{0.7em}57.73 		&\hspace{0.7em}59.00 		&\hspace{0.7em}50.68 		&\hspace{0.7em}68.12 		&\hspace{0.7em}66.72 		&\hspace{0.7em}66.46 		&\hspace{0.7em}65.09         \\
\rowcolor{mygreen} Delta                 &\hspace{0.7em}-0.06 		&\hspace{0.7em}-0.07 		&\hspace{0.7em}+0.03 		&\hspace{0.7em}+0.03		&\hspace{0.7em}\textbf{+2.17} 		&\hspace{0.7em}\textbf{+1.77} 		&\hspace{0.7em}\textbf{+0.82} 		&\hspace{0.7em}\textbf{+0.70} 		&\hspace{0.7em}+0.13 		&\hspace{0.7em}+0.12 		&\hspace{0.7em}\textbf{+0.48} 		&\hspace{0.7em}\textbf{+0.45}
\\
\bottomrule[0.8pt]
\end{tabularx}%}
\end{table*}

\begin{table*}[]
    \centering
    % \fontsize{5pt}{\baselineskip}\selectfont%设置字体大小
    \caption{Comparative experiments on the KITTI 3d object detection validation benchmark. All methods are based on PointPillars.}
    \label{table_10}
	\renewcommand\arraystretch{0.95}%行高
	% \resizebox{\linewidth}{!}{
\begin{tabularx}{\textwidth}{c|XXXXXXXXX|c}
\specialrule{0.8pt}{0pt}{0pt}%第一行线宽
\multirow{2}{*}{Method}& \multicolumn{3}{@{\hspace{1.1em}}c}{Car} & \multicolumn{3}{@{\hspace{0.2em}}c}{Pedestrian} & \multicolumn{3}{@{\hspace{-0.3em}}c|}{Cyclist} & \multirow{2}{*}{Times(ms)} \\
                        & \hspace{1.3em}Easy      & \hspace{1.2em}Mod.       & \hspace{1.1em}Hard     & \hspace{1.0em}Easy        & \hspace{0.9em}Mod.         & \hspace{0.8em}Hard        & \hspace{0.7em}Easy       & \hspace{0.6em}Mod.        & \hspace{0.5em}Hard           \\ \hline
Traditional-NMS                 &\hspace{1.3em}87.75 	  &\hspace{1.2em}78.40 	  &\hspace{1.1em}75.18 	  &\hspace{1.0em}57.30 	  &\hspace{0.9em}51.41 	  &\hspace{0.8em}46.87 	  &\hspace{0.7em}81.57 	  &\hspace{0.6em}62.81 	  &\hspace{0.5em}58.83     & \textcolor{black}{\textbf{6.30}} \\
Soft-NMS                 &\hspace{1.3em}77.35 	  &\hspace{1.2em}68.78 	  &\hspace{1.1em}63.76 	  &\hspace{1.0em}28.07 	  &\hspace{0.9em}25.07 	  &\hspace{0.8em}21.23 	  &\hspace{0.7em}60.87 	  &\hspace{0.6em}44.24 	  &\hspace{0.5em}39.94     & \textcolor{black}{7.83} \\
DIoU-NMS                 &\hspace{1.3em}87.68 	  &\hspace{1.2em}78.31 	  &\hspace{1.1em}75.08 	  &\hspace{1.0em}59.00 	  &\hspace{0.9em}52.79 	  &\hspace{0.8em}48.06 	  &\hspace{0.7em}\textbf{82.15} 	  &\hspace{0.6em}63.01 	  &\hspace{0.5em}58.97     & \textcolor{black}{17.88} \\
\rowcolor{mygreen} Fuzzy-NMS                &\hspace{1.3em}\textbf{88.09} 	  &\hspace{1.2em}\textbf{78.77} 	  &\hspace{1.1em}\textbf{75.45} 	  &\hspace{1.0em}\textbf{59.45} 	  &\hspace{0.9em}\textbf{53.62} 	  &\hspace{0.8em}\textbf{49.06} 	  &\hspace{0.7em}81.73 	  &\hspace{0.6em}\textbf{63.17} 	  &\hspace{0.5em}\textbf{59.50}     & \textcolor{black}{10.08} \\ \bottomrule[0.8pt]
\end{tabularx}%}
\end{table*}

%tabel
\begin{table*}[]
	\centering
    % \fontsize{5pt}{\baselineskip}\selectfont%设置字体大小
	\caption{Comparative experiments on the KITTI BEV object detection validation benchmark. All methods are based on PointPillars.}
    \label{table_11} 
	\renewcommand\arraystretch{0.95}%行高
	% \resizebox{\linewidth}{!}{
\begin{tabularx}{\textwidth}{c|XXXXXXXXX|c}
\specialrule{0.8pt}{0pt}{0pt}%第一行线宽
\multirow{2}{*}{Method}& \multicolumn{3}{@{\hspace{1.1em}}c}{Car} & \multicolumn{3}{@{\hspace{0.2em}}c}{Pedestrian} & \multicolumn{3}{@{\hspace{-0.3em}}c|}{Cyclist} & \multirow{2}{*}{Times(ms)} \\
                        & \hspace{1.3em}Easy      & \hspace{1.2em}Mod.       & \hspace{1.1em}Hard     & \hspace{1.0em}Easy        & \hspace{0.9em}Mod.         & \hspace{0.8em}Hard        & \hspace{0.7em}Easy       & \hspace{0.6em}Mod.        & \hspace{0.5em}Hard           \\ \hline
Traditional-NMS                &\hspace{1.3em}92.04 	   &\hspace{1.2em}88.06 	   &\hspace{1.1em}\textbf{86.66} 	   &\hspace{1.0em}61.60 	   &\hspace{0.9em}56.01 	   &\hspace{0.8em}52.05 	   &\hspace{0.7em}85.26 	   &\hspace{0.6em}66.24 	   &\hspace{0.5em}62.22     & \textcolor{black}{\textbf{6.30}} \\
Soft-NMS                &\hspace{1.3em}86.69 	   &\hspace{1.2em}81.60 	   &\hspace{1.1em}75.46 	   &\hspace{1.0em}31.10 	   &\hspace{0.9em}26.19 	   &\hspace{0.8em}22.15 	   &\hspace{0.7em}64.94 	   &\hspace{0.6em}47.60 	   &\hspace{0.5em}43.17     & \textcolor{black}{7.83} \\
DIoU-NMS                &\hspace{1.3em}91.96 	   &\hspace{1.2em}87.95 	   &\hspace{1.1em}86.55 	   &\hspace{1.0em}63.25 	   &\hspace{0.9em}57.68 	   &\hspace{0.8em}53.42 	   &\hspace{0.7em}\textbf{85.92} 	   &\hspace{0.6em}66.35 	   &\hspace{0.5em}62.43     & \textcolor{black}{17.88} \\
\rowcolor{mygreen} Fuzzy-NMS               &\hspace{1.3em}\textbf{92.45} 	   &\hspace{1.2em}\textbf{88.48} 	   &\hspace{1.1em}85.80 	   &\hspace{1.0em}\textbf{64.11} 	   &\hspace{0.9em}\textbf{58.37} 	   &\hspace{0.8em}\textbf{54.39} 	   &\hspace{0.7em}85.87 	   &\hspace{0.6em}\textbf{66.78} 	   &\hspace{0.5em}\textbf{62.98}     & \textcolor{black}{10.08} \\ \bottomrule[0.8pt]
\end{tabularx}%}
\end{table*}

\begin{figure*}[!t]
	\vspace{-5mm}
	\centering
	\includegraphics[width=0.97\linewidth]{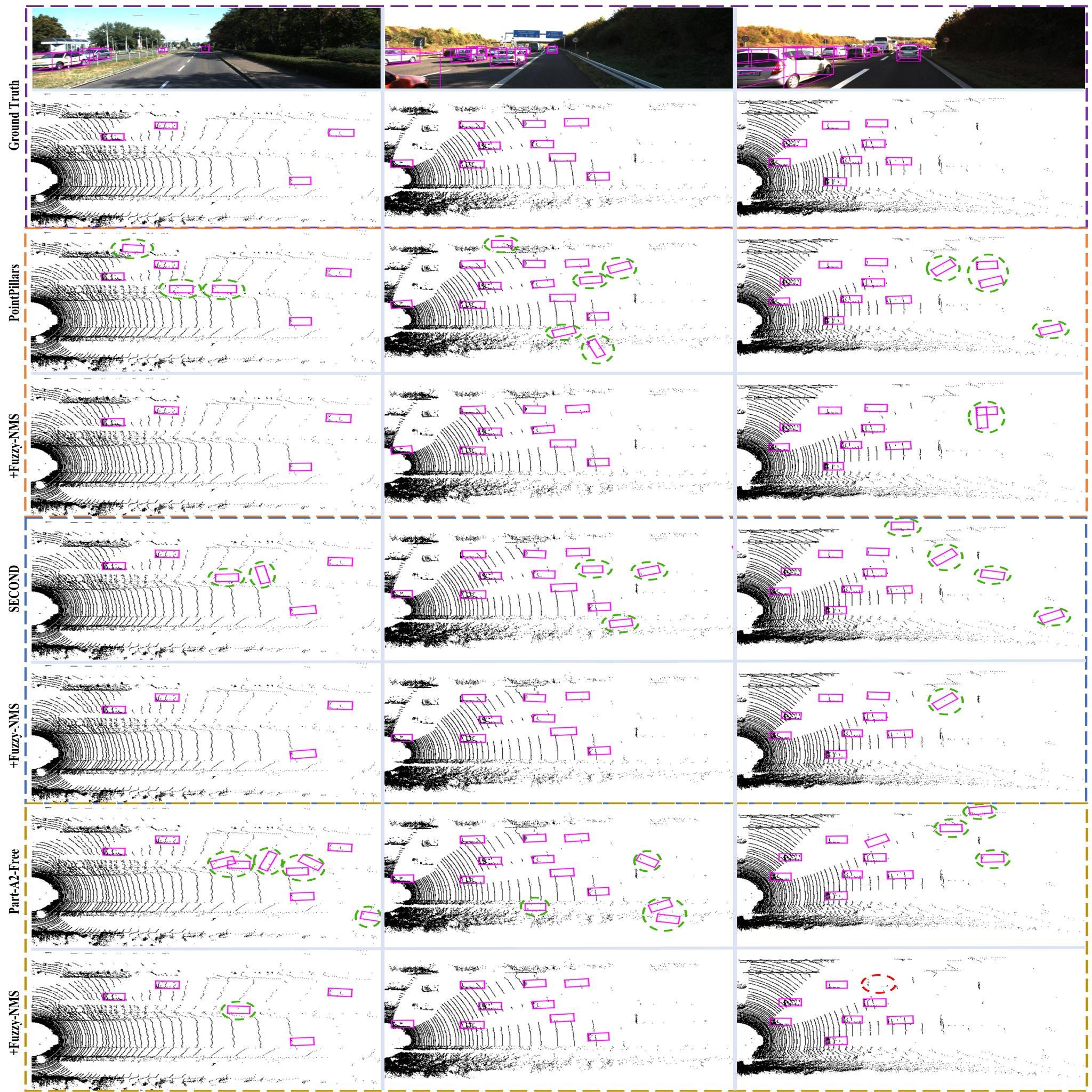}
	\caption{Qualitative BEV visualization results for ordinary highway and expressway scenes. The eight rows represent ground truth in images (GT), ground truth in BEV point cloud, PointPillars, PointPillars + Fuzzy-NMS, SECOND, SECOND + Fuzzy-NMS, Part-A2-Free, and Part-A2-Free + Fuzzy-NMS, respectively. Purple rectangular boxes show detection results, green dotted boxes display false detection results, and dotted boxes indicate missed detection results. From the experimental results, a lot of false detection boxes are removed when adding our proposed Fuzzy-NMS module in these baselines.
		\vspace{-4mm}}
	\label{fig9}
\end{figure*}

To better illustrate this comparison, the moderate results of these five baseline methods without and with our proposed Fuzzy-NMS module were visualized using a histogram, as shown in \mbox{Fig. \ref{fig8}}. In the figure, three categories, including car, pedestrian, and cyclist, were given, respectively. And the results of the baseline methods were represented by light-colored columns, while the methods with our Fuzzy-NMS module corresponded to the darker-colored columns. It was evident the performance of the five baseline models had been improved to varying degrees after the addition of our module. It was also worth noting the Fuzzy-NMS module primarily displayed improvements for pedestrians and cyclists (small objects), which was consistent with the original intention of designing a Fuzzy-NMS. 

The effects of this algorithm were further assessed by conducting qualitative visualization experiments, as shown in \mbox{Fig. \ref{fig9}}. We visualized these outcomes in BEV for PointPillars, SECOND, and Part-A2-Free models in ordinary highway and expressway scenes to compare intuitively. The purple rectangular boxes denoted detection results, the green dotted boxes were false detection results from the baseline, and the red dotted boxes represented missed detection results after the addition of the Fuzzy-NMS module. This algorithm helped eliminate some false detection, especially for distant sparse objects. It was worth emphasizing that each of the five models used the same parameters. In other words, the Fuzzy-NMS module, which improved the performance of PointPillars, could also improve other models (i.e., SECOND). The proposed method is a general plug-and-play module, and the resulting detection accuracy can be improved without further parameter adjustment for a given data set. In addition, Fuzzy-NMS was also compared with other famous NMS methods, including traditional-NMS, Soft-NMS \cite{bodla2017soft}, and DIoU-NMS \cite{2019Distance}, to prove the performance. The results of a 3D comparison were shown in \mbox{Table \ref{table_10}}, while BEV results were provided in \mbox{Table \ref{table_11}}. PointPillars was utilized as a baseline in these experiments. The proposed algorithm achieved 3D mAP values of 78.77$\%$, 53.62$\%$, and 63.17$\%$, and BEV mAP values of 88.48$\%$, 58.37$\%$, and 66.78$\%$ for car, pedestrian, and cyclist categories at a moderate level, respectively. The cumulative performance of Fuzzy-NMS was superior to each of the other NMS modules. 
To improve the real-time performance of the algorithm, we performed code acceleration by using C++ language instead of Python, which was presented in the form of a dynamic library. \textcolor{black}{Finally, the run-time was about 10.08ms, which was inferior to that of conventional NMS and Soft-NMS but superior to DIoU-NMS. We also record the running time of the detector. When the Fuzzy-NMS module is not added, the complete detection process of the original PointPillars detector takes 48.10ms per frame, and the total time spent after adding the Fuzzy-NMS module is 55.38ms per frame.} These results suggest that Fuzzy-NMS is a competitive post-processing model for 3D object detection tasks.

\section{CONCLUSION}
In this study, a novel Fuzzy-NMS module is proposed to solve the problem of extensive threshold filtering in 3D object detection tasks. This method utilizes the density and volume of candidate boxes to divide them into LVHD, SVHD, and LD categories, allocating optimal thresholds for each to improve the filtering capabilities of NMS further. Adequate experiments involving the KITTI \textcolor{black}{and Waymo datasets} prove\textcolor{black}{s} that the addition of the proposed module can produce significant improvements on 3D and BEV mAP in \textcolor{black}{many benchmarks.} This outcome demonstrates that Fuzzy-NMS can remarkably improve the accuracy of 3D object detection and, as a generalized plug-and-play module, can be directly added to a 3D detection framework to improve performance without the need for training. \textcolor{black}{In addition, works like 
\cite{shen2021distilled}and \cite{zhao2021real} show that light weighted networks will definitely occupy a very important place in the development of autonomous driving in the future. We will further promote the development of lightweight networks and NMS in the field of object detection, which will provide a more efficient and accurate object detection solution for devices with constrained computing resources, and have wide-ranging implications in practical applications.}

\section{Acknowledgments}
We thank LetPub (www.letpub.com) for linguistic assistance and pre-submission expert review.

\bibliographystyle{Transactions-Bibliography/IEEEtran}
\bibliography{Transactions-Bibliography/IEEEabrv,Transactions-Bibliography/BIB_xxx}\ %IEEEabrv instead of IEEEfull

\end{CJK}
\end{document}